\documentclass[pdflatex,sn-mathphys-num]{sn-jnl}

\usepackage{graphicx}%
\usepackage{multirow}%
\usepackage{amsmath,amssymb,amsfonts}%
\usepackage{amsthm}%
\usepackage{mathrsfs}%
\usepackage[title]{appendix}%
\usepackage{xcolor}%
\usepackage{textcomp}%
\usepackage{manyfoot}%
\usepackage{booktabs}%
\usepackage{algorithm}%
\usepackage{algorithmicx}%
\usepackage{algpseudocode}%
\usepackage{listings}%
\usepackage{hyperref}
\usepackage{url}
\usepackage{color}
\usepackage{siunitx}
\usepackage{graphicx}
\usepackage{wrapfig}
\usepackage{caption}
\usepackage{subcaption}
\usepackage{booktabs}
\usepackage{glossaries}
\usepackage{rotating}
\usepackage{pifont}

\graphicspath{ {./} }

\theoremstyle{thmstyleone}%
\theoremstyle{thmstyletwo}%

\theoremstyle{thmstylethree}%

\begin{document}

\title[Don't Get Me Wrong]{Don't Get Me Wrong: How to Apply Deep Visual Interpretations to Time Series}



%
%
%

\author*[1]{\fnm{Christoffer} \sur{L\"offler}\textsuperscript{\textdagger}\textsuperscript{\textdaggerdbl}}\email{christoffer.loffler@pucv.cl} 

\author[4]{\fnm{Wei-Cheng}
\sur{Lai}\textsuperscript{\textdagger}\email{wei-cheng.lai@hpi.de}} 

\author[2]{\fnm{Dario} \sur{Zanca}}\email{dario.zanca@fau.de}

\author[3]{\fnm{Lukas} \sur{Schmidt}\textsuperscript{\textdaggerdbl}\email{lukas.schmidt.er@gmail.com}}

\author[2]{\fnm{Bj\"orn M.}
\sur{Eskofier}\email{bjoern.eskofier@fau.de}}

\author[3]{\fnm{Christopher} \sur{Mutschler}}\email{christopher.mutschler@iis.fraunhofer.de}

\affil[1]{\orgdiv{School of Computer Engineering}, \orgname{Pontificia Universidad Católica de Valparaíso}, \orgaddress{\street{Brasil 2950}, \city{Valparaíso}, \postcode{2340025}, \country{Chile}}}

\affil[2]{\orgdiv{Machine Learning and Data Analytics Lab}, \orgname{Friedrich-Alexander University Erlangen-N\"urnberg}, \orgaddress{\street{Carl-Thiersch-Straße 2b}, \city{Erlangen}, \postcode{91052}, \country{Germany}}}

\affil[3]{\orgdiv{Positioning and Networks Division}, \orgname{Fraunhofer Institute for Integrated Circuits IIS}, \orgaddress{\street{Nordostpark 93}, \city{Nuremberg}, \postcode{90411}, \country{Germany}}}

\affil[4]{\orgdiv{Digital Health and Machine Learning}, \orgname{Hasso-Plattner-Institute}, \orgaddress{\street{Prof.-Dr.-Helmert Straße 2-3}, \city{Potsdam}, \postcode{14482}, \country{Germany}}}

\renewcommand{\thefootnote}{\relax} 
\footnotetext{\textsuperscript{\textdagger} With Friedrich-Alexander University Erlangen-Nuremberg when conducting this research.}
\footnotetext{\textsuperscript{\textdaggerdbl} With Fraunhofer Institute for Integrated Circuits IIS when conducting this research.}
\setcounter{footnote}{0}
\renewcommand{\thefootnote}{\arabic{footnote}} 


\abstract{The correct interpretation of convolutional models is a hard problem for time series data. While saliency methods promise visual validation of predictions for image and language processing, they fall short when applied to time series. These tend to be less intuitive and represent highly diverse data, such as the tool-use time series dataset. Furthermore, saliency methods often generate varied, conflicting explanations, complicating the reliability of these methods. Consequently, a rigorous objective assessment is necessary to establish trust in them.
This paper investigates saliency methods on time series data to formulate recommendations for interpreting convolutional models and implements them on the tool-use time series problem. To achieve this, we first employ nine gradient-, propagation-, or perturbation-based post-hoc saliency methods across six varied and complex real-world datasets. Next, we evaluate these methods using five independent metrics to generate recommendations. Subsequently, we implement a case study focusing on tool-use time series using convolutional classification models. 
Our results validate our recommendations that indicate that none of the saliency methods consistently outperforms others on all metrics, while some are sometimes ahead. Our insights and step-by-step guidelines allow experts to choose suitable saliency methods for a given model and dataset.}



\maketitle

\section{Introduction}\label{sec: introduction}

Deep learning (DL) has become increasingly popular in many real-world applications that process multi-modal time series data~\citep{fawaz_tsc}.
While we fundamentally rely on the networks' correct operation in many applications that consider safety~\citep{berkenkamp2017safe}, they remain difficult to interpret. Typical applications are fault detection in industry~\cite{huangDeepRootCause2024} and monitoring of industrial processes~\citep{10.1007/978-3-030-67670-4_33_embedded}, support of healthcare~\citep{iqbalAutomatedIdentificationHuman2022} and sports~\citep{faucris.240294847}, or safety in autonomous driving~\citep{Schmidt2021}. The need for improved model understanding~\citep{Carvalho_2019}, along with regulatory guidelines~\citep{goodman2017european}, led to a myriad of new approaches to the visual interpretation problem~\citep{zhang2018visual}.
 
Post-hoc visual interpretation allows a domain expert to validate and understand how (almost) arbitrary deep learning models operate. Their central idea is highlighting features on the input that are "relevant" for the prediction of a learned model~\citep{adebayoSanityChecksSaliency2018}. Many of these techniques do not require a modification of the original model~\citep{Simonyan2014DeepIC, 10.1145/2939672.2939778}, which makes them compatible with different architectures. Thus, they are helpful as a general-purpose validation tool for neural networks across different tasks~\citep{arrieta2020explainable}.

However, while visual interpretation yields intuitive and correct explanations for images~\citep{XDL2021samek}, the application of these methods on time series data is still a generally unsolved problem~\citep{rojat2021explainable}. Time series are inherently more diverse~\citep{rojat2021explainable} because they may originate from various sensors and processes and often do not allow a prominent patch- or texture-based localization of critical features for human observers. Even though success is possible, e.g., when analyzing individual gait patterns~\cite{horstExplainingUniqueNature2019}, applying and evaluating visual interpretability methods is difficult. 

This paper investigates automated quality assurance of industrial processes that use hand-held tools. We equip the tools with a sensor platform that records multi-modal sensor readings, including an accelerometer, a gyroscope, and a magnetic field sensor. These data allow the detailed prediction of work steps in an assembly process involving, among other tools, electric screw-drivers or rivet guns\footnote{Data available under \url{https://github.com/mutschcr/tool-tracking}}~\citep{10.1007/978-3-030-67670-4_33_embedded, lofflerAutomatedQualityAssurance2021}. 
In this industrial setting, even domain experts cannot easily judge if explanations are correct in (i) delivering the reasoning of the decision process in the DL model and (ii) capturing the actual features in the dataset that lead to a correct classification.

Hence, it is essential not to apply different visualization methods mindlessly. This requires reliable, quantitative quality metrics that evaluate visual interpretations and enable an expert to select a suitable visualization for a given model and data. However, both state-of-the-art visualization techniques and metrics that assess visual interpretations, e.g., Pixel Flipping~\citep{pixelflipping}, Sanity Check~\citep{adebayoSanityChecksSaliency2018}, and sensitivity checks~\citep{rebuffi2020saliency}, so far have been examined on images~\citep{rojat2021explainable} or NLP tasks~\citep{arras-etal-2017-explaining_lrp_lstm}, but only rarely on more diverse time series~\cite{Schlegel2019TowardsAR}. Furthermore, studies on the quality of saliency maps focus on only one quality metric~\citep{Schlegel2019TowardsAR,turbeEvaluationPosthocInterpretability2023,nguyenRobustExplainerRecommendation2024}, which limits their reliability~\cite{tomsettSanityChecksSaliency2020,hedstromQuantusExplainableAI2023}. This lack of thorough evaluation seriously limits the application and utility of saliency methods for time series classification.

This paper investigates the evaluation of saliency methods on complex time series data and formulates recommendations for interpreting the explanations on the example of tool-use time series. 
\begin{itemize}
    \item We train Fully Convolutional Networks (FCN)~\citep{long2015fully} and Temporal Convolutional Networks (TCN)~\citep{bai2018empirical} on different classification datasets from the UCR repository~\citep{dau18archive} (GunPointAgeSpan, FordA, FordB, ElectricDevices, MelbournePedestrian, NATOPS). 
    \item We apply nine gradient-, propagation, or perturbation-based post-hoc saliency methods: Gradient~\citep{Simonyan2014DeepIC}, Integrated Gradient~\citep{sundararajan2017axiomatic}, SmoothGrad~\citep{smilkov2017smoothgrad}, Guided Backpropagation~\citep{Springenberg2015StrivingFS}, GradCAM~\citep{8237336}, Guided-GradCAM~\citep{8237336}, Layer-Wise Relevance Propagation (LRP)~\citep{10.1371/journal.pone.0130140}, LIME~\citep{10.1145/2939672.2939778}, and Kernel SHAP~\citep{NIPS2017_8a20a862}.
    \item We rate the methods along five orthogonal metrics to rate and validate distinct qualities of saliency methods: "sanity"~\citep{adebayoSanityChecksSaliency2018}, "faithfulness"~\citep{melisRobustInterpretabilitySelfExplaining2018, Schlegel2019TowardsAR}, "sensitivity"~\citep{rebuffi2020saliency}, "robustness"~\citep{yehFidelitySensitivityExplanations2019}, "stability"~\citep{fel2020representativity,li2021quantitative}.
    \item We give clear recommendations that allow experts to choose and validate appropriate saliency methods for time series data and implement them on the multi-class, multi-modal tool-use time series problem~\citep{10.1007/978-3-030-67670-4_33_embedded, lofflerAutomatedQualityAssurance2021}.
\end{itemize}

The remainder of this article is structured as follows. Section~\ref{sec: preliminaries} formulates the problem. 
Section~\ref{sec: methodology} presents scoring categories of saliency metrics.
Section~\ref{sec: experimental setup} discusses the experimental setup, including the choices of models, datasets, and saliency methods, and Section~\ref{sec: experiments} presents the results, derives recommendations, and demonstrates their implementation for the tool-use problem. Section~\ref{sec: discussion} discusses the contributions and their limitations, and Section~\ref{sec: conclusion} concludes.

\section{Problem Formulation}
\label{sec: preliminaries}

\textbf{Classification task.} A multivariate time series is defined by $X = [X^1, ..., X^H]$, where $H$ is the number of input channels,  $X^i = (x_1^i, ..., x_T^i) \in \mathbb{R}^T$ is an ordered set of real values, and $T$ denotes the number of time steps. For $H=1$, we consider a univariate time series; otherwise, we consider a multivariate time series. Time series often include complex temporal dependencies, i.e., distinct points are not independent. Time series classification defines a mapping $X \rightarrow y$ (the model $m$) that minimizes the error on a dataset $D = \{(X_1, y_1), ..., (X_N, y_N)\}$, where $N$ is the number of data samples, $X \in D$ is a time series, $y_i \in \mathbb{R}^{C}$ denotes the one-hot vector of a class label the input belongs to, and $C$ is the number of classes. 


\textbf{Saliency Methods.} We divide methods that generate saliency for neural networks into two types, ante-hoc methods, which are inherently part of the model, and post-hoc methods, which provide the interpretation after training~\citep{rojat2021explainable}. We focus on post-hoc techniques and further divide them into i) gradient-, ii) propagation-, and iii) perturbation-based methods~\citep{li2021quantitative,warnecke2020evaluating_secure,NEURIPS2020_47a3893c_ts_interpret,letzgusExplainableArtificialIntelligence2022}. 
Post-hoc visual interpretation methods compute a relevance map $M_m^c \in \mathbb{R}^{H \times T}$, $M_m^c(X) = [R^1, ..., R^H]$, where $R^i = (r_1^i, ..., r_T^i)$, representing the importance of each input feature wrt. the class $c$ and a model $m$, for each time step. We use $M$ as a function to produce the saliency map. For clarity, we will omit the dependency on the model $m$, i.e., $M_m^c \equiv M^c$, if it is not explicitly required.

\begin{figure}[t!]
    \centering
    \begin{subfigure}[b]{0.3\textwidth}
    \centering\includegraphics[width=1.\textwidth, keepaspectratio]{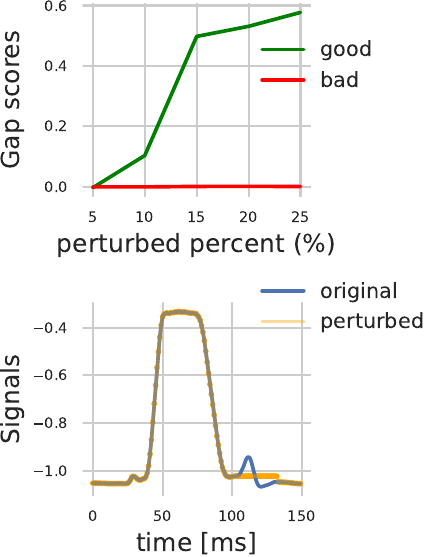}
         \caption{Faithfulness}
         \label{fig 1: faith}
    \end{subfigure}
    \begin{subfigure}[b]{0.3\textwidth}
         \centering
         \includegraphics[width=1.\textwidth, keepaspectratio]{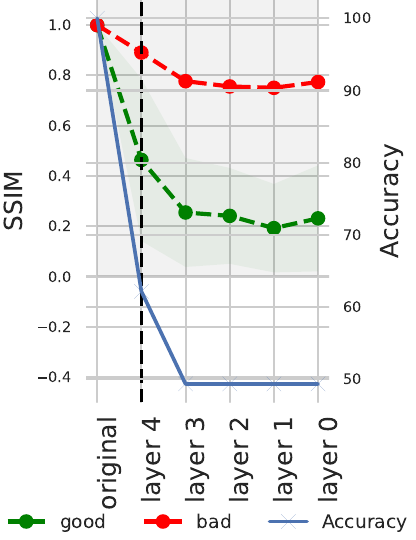}
         \caption{Sanity}
         \label{fig 1: sane}
    \end{subfigure}
    \begin{subfigure}[b]{0.3\textwidth}
         \centering
         \includegraphics[width=1.\textwidth, keepaspectratio]{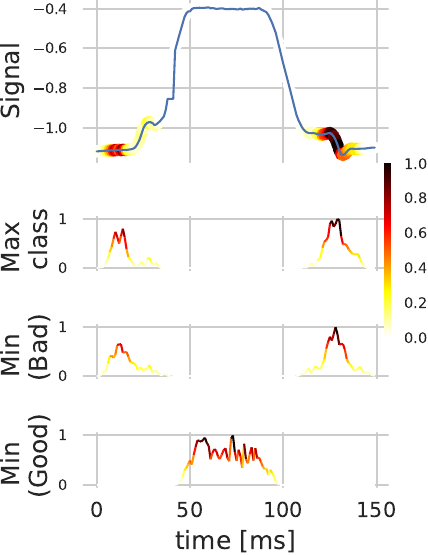}
         \caption{Sensitivity}
         \label{fig 1: sensible}
    \end{subfigure}
    
    \begin{subfigure}[b]{0.3\textwidth}
         \centering
         \includegraphics[width=1.\textwidth, keepaspectratio]{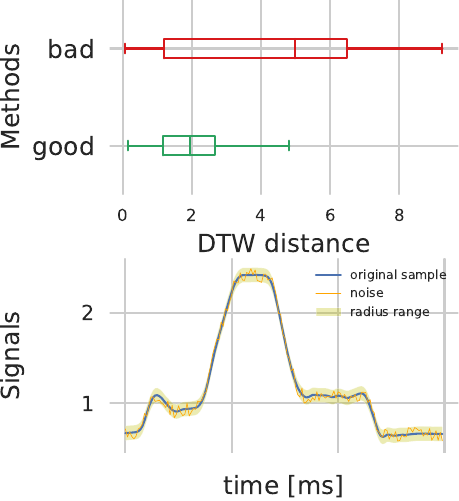}
         \caption{Robustness}
         \label{fig 1: robust}
    \end{subfigure}
    \begin{subfigure}[b]{0.3\textwidth}
         \centering
         \includegraphics[width=1.\textwidth, keepaspectratio]{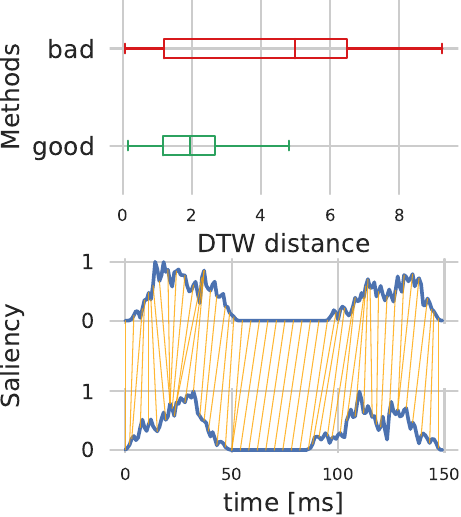}
         \caption{Stability}
         \label{fig 1: stable}
    \end{subfigure}
  \caption{We show examples of a good (green) and a bad (red) score for each metric: (\subref{fig 1: faith}) faithful saliency correlates with predictive accuracy, tested by perturbing the input sequence;
  (\subref{fig 1: sane}) sane saliency depends on network parameters, tested by randomizing weights and biases; (\subref{fig 1: sensible}) sensitive saliency of the predicted class in one sample is different from others; (\subref{fig 1: robust}) for robust saliency small changes to input data cause only small effects; (\subref{fig 1: stable}) stable saliency assigns relevance to similar features for the same class.}
  \label{fig 1}%
\end{figure}

i) Gradient-based methods compute the relevance for all input features by passing gradients backward through the neural network~\citep{baehrensHowExplainIndividual2010}. Gradient~\citep{Simonyan2014DeepIC} computes the saliency map $M^c$ of a class $c$ using the derivative of the class score $P^c$ of the model concerning the input sample $x$, as $M^c(x) = \frac{\partial P^c}{\partial x}$. The gradient indicates the importance of points in the input sequence for predictions. 
The advantage of gradient-based methods lies in their computational efficiency, as they use only a small number of backward passes to compute $M^c(x)$.

ii) Propagation-based methods leverage the network's structure to produce the explanation~\citep{letzgusExplainableArtificialIntelligence2022}. They propagate the output predictions back to the input. Methods such as Layerwise Relevance Propagation (LRP)~\citep{10.1371/journal.pone.0130140} implement propagation rules for different layer types. LRP computes relevance $R_j$ at layer $j$ based on the contributions from the relevance $R_k$ of the consecutive layer at index $k$. Let $Z_{jk}$ denote the forward pass contribution of neuron $j$ to neuron $k$. LRP then sums over previous contributions $\sum_{0,j}$ and current contributions $\sum_{k}$ to get the relevancy $R_j = \sum_k{\frac{Z_{jk}}{\sum{Z_jk}_{0,j}} R_k}$. We refer to Letzgus et al.~\citep{letzgusExplainableArtificialIntelligence2022}.

iii) Perturbation-based methods perturb known input samples and measure the effects of specific perturbations on the predicted class via a forward pass through the network. For instance, Local Interpretable Model-Agnostic Explanations (LIME)~\citep{10.1145/2939672.2939778} fits a local surrogate model (e.g., a linear regression) as an explanation and calculates relevance based on this surrogate. 
Perturbation-based methods are computationally expensive as they require multiple forward passes per sample. However, they do not need gradient information and work with black-box models.

\textbf{Saliency Metrics.}
Saliency metrics aim to provide a statistical measure that captures actual model behavior and does not depend on human feedback. They aggregate performance metrics automatically~\citep{rojat2021explainable}, and use proxy tasks to generate quantitative evaluations. An evaluation metric for saliency methods defines a score $\mathcal{S_{\text{metric}}}(\cdot)$ that rates the quality of the relevance map $M$ at a sample $X$ given a model $m$ and optional parameters. We provide a unified view so that for all scores, a higher score corresponds to a better visualization according to the perspective.

Even with a set of saliency metrics, no saliency method dominates the others on all models and datasets~\cite{tomsettSanityChecksSaliency2020,liExperimentalStudyQuantitative2021}. This article proposes a metrics framework with orthogonal categories, specifically for time series, and adapts and extends saliency metrics to (multivariate) time series. Then, it demonstrates its implementation on the tool-use problem.

\section{Scoring Categories}
\label{sec: methodology}

We propose five distinct categories ("sanity", "faithfulness", "sensitivity", "robustness", and "stability") to assess saliency methods and to determine their performance and trustworthiness in classification tasks on time series. For each of them, we propose a metric that enables a comparative evaluation of diverse saliency methods.

\begin{figure}
    \centering
    \includegraphics[width=0.50\textwidth,keepaspectratio,
    trim=1.5cm 0.4cm 2.5cm 0.5cm,
    clip]{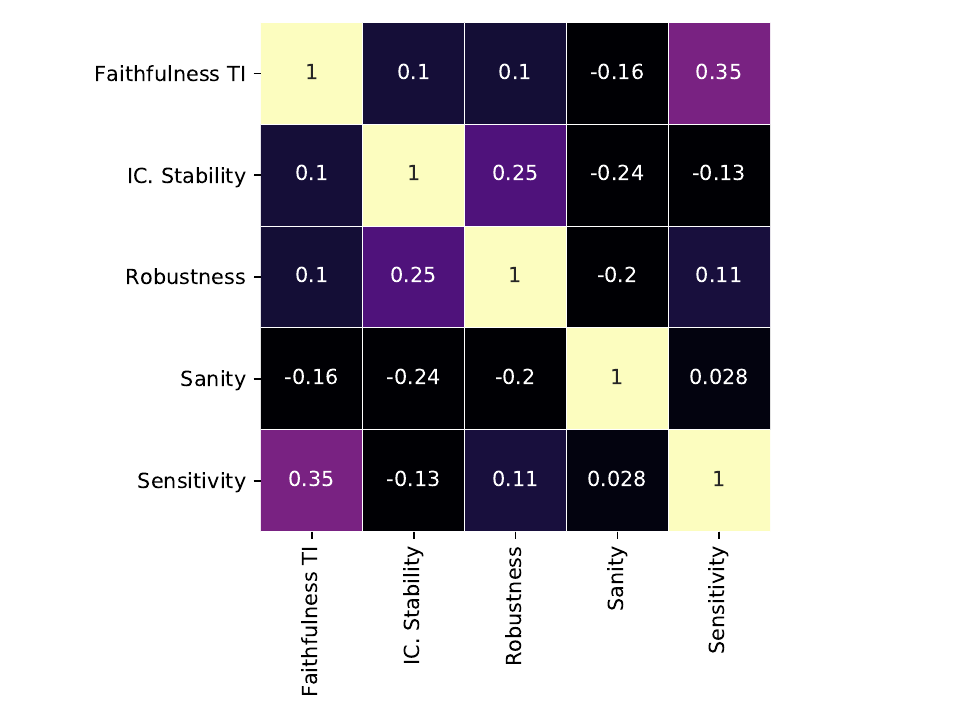} 
    \caption{Pairwise Pearson correlations of scores. Every metric is independent of all others.}
    \label{results: correlations_heatmap}
\end{figure}

\textbf{Why do we need five scoring categories?} It seems tempting to rely on a single metric or aggregated score across multiple metrics to assess the quality of a visual interpretation. However, we show that our five presented categories are orthogonal (see Fig.~\ref{results: correlations_heatmap}) and capture distinct attributes. This finding also confirms previously discussed studies~\cite{liExperimentalStudyQuantitative2021,tomsettSanityChecksSaliency2020}. It extends work by Hedstr\"om et al.~\citep{hedstromQuantusExplainableAI2023}, who loosely cluster quantitative evaluation metrics into six groups based on their intuitive similarity~\footnote{parallel work, see our preprint \url{https://arxiv.org/abs/2203.07861}}. Interpretations depend on model parameters (sanity check~\citep{adebayoSanityChecksSaliency2018}), predictive features (faithfulness~\citep{melisRobustInterpretabilitySelfExplaining2018}), coherence of class predictions (intra-class stability~\citep{fel2020representativity}), robustness against noise (max sensitivity with adversarial noise~\citep{yehFidelitySensitivityExplanations2019}), specific sequences (or even points) in a time series like by \cite{NEURIPS2018_pr_re}, such as the relevance map's specificity (inter-class sensitivity~\citep{rebuffi2020saliency}). It is crucial to assess if a given interpretation accurately captures these dependencies.

We briefly overview our metrics in Fig.~\ref{fig 1}. As a set, they allow in-depth analysis of saliency methods' performance on time series data while still being independent, see also Sec.~\ref{sec: recommendations} for details. %

\begin{itemize}
\vspace*{-\topsep}
    \item[(\subref{fig 1: faith})] \textit{faithful saliency} correlates with predictive accuracy. Perturbing a percentage of the input sequence with high saliency decreases accuracy~\citep{melisRobustInterpretabilitySelfExplaining2018}. 
    \item[(\subref{fig 1: sane})] \textit{sane saliency} depends on network parameters and is structurally different after randomizing the network's weights $\rho_i$ for layers $[1,2,3]$~\citep{adebayoSanityChecksSaliency2018}. 

    \item[(\subref{fig 1: sensible})]   saliency is \textit{sensitive} if the predicted (max) class in one sample is sufficiently different from any other (min) class~\citep{rebuffi2020saliency}. 
    \item[(\subref{fig 1: robust})]  saliency is \textit{robust} if small changes to the input cause only small changes to the saliency~\citep{yehFidelitySensitivityExplanations2019}. 
     \item[(\subref{fig 1: stable})]  saliency is \textit{stable} if it assigns relevance to similar features for all samples of a class, wrt. a suitable distance metric~\citep{fel2020representativity}.
\end{itemize}

\subsection{Faithfulness}
\label{sec: faithfulness}

A relevance measure is faithful if input features with a high relevance (w.r.t. $M^c$) have a high influence on the model prediction, see Fig.~\ref{fig 1: faith} for an illustration. \cite{melisRobustInterpretabilitySelfExplaining2018} propose a perturbation-based metric that evaluates the faithfulness of predictions. The metric measures the correlation between input (features) with high saliency on the one hand and predictive accuracy on the other hand. For time series data, two variants of faithfulness are based on different data distribution assumptions. Both rank inputs of the time step of $X_T \in \mathbb{R^T}$ according to the relevance of its saliency map. However, they differ in perturbations: \cite{melisRobustInterpretabilitySelfExplaining2018} (further called Temporal Importance (TI)) strictly adheres to the ranked importance of saliency to choose what to perturb.
In contrast, \cite{Schlegel2019TowardsAR} assume temporal correlation of inputs and perturb a connected sub-sequence instead (Temporal Sub-sequences (TS)). While the latter is a more intuitive model for perturbations on time series data, it may misrepresent the importance of features, e.g., if they are not clustered or do not follow the assumption that they should be sequential. Furthermore, there may be several crucial sub-sequences in a single sample.

We use either variant of faithfulness on multivariate data $X^{H}_{T}$ as follows. Based on the relevance $r$, we select time points $t \in \{1, ..., T\}$ and features $i \in \{1, ..., H\}$.
For TI, we select and perturb the inputs of the sample $X$ with the highest relevance $r_{t}^{i}$ and replace their values with the dataset mean at time $t$, as proposed by~\cite{melisRobustInterpretabilitySelfExplaining2018}.
For TS, we select $t$ according to the maximum relevance $r_{t}^{i}$ and perturb a sub-sequence of the inputs of the sample $X$ by replacing its values also with a mean value of the data~\cite{melisRobustInterpretabilitySelfExplaining2018} of length $L$ at time $t$, as $x_{\text{mean}} = (x_{t+\frac{L}{2}}^{i}, ..., x_{t}^{i}, ..., x_{t-\frac{L}{2}}^{i})$, to break the temporal correlation, following~\cite{Schlegel2019TowardsAR}.
We choose to perturb samples with mean values, making it less likely to move them off their manifold~\cite{montavonMethodsInterpretingUnderstanding2018}.

The perturbed sample is denoted as $X'$. We compute the mean faithfulness scores over the whole dataset $D$ for TI as
\begin{equation}\textstyle
    S_{\text{Faithfulness TI}}(M, D, m) = 
    - \frac{1}{N \cdot L} \sum_{X \in D} \sum_{l=0}^{L} m^c(X'_l)
\end{equation}
and for TS, as
\begin{equation}\textstyle
    S_{\text{Faithfulness TS}}(M, D, m) =  \frac{1}{N} \cdot \sum_{X \in D}m^c(X) - m^c(X'),
\end{equation}
where $m^c: R^c \rightarrow R$ is the softmax prediction of the target class $c$. For TS, the gap score between the softmax prediction of the original and the perturbed sample is $m^c(X) - m^c(X')$. For TI, we use the Area Under Curve (AUC) instead, where $L$ is the total perturbation length and $X'_l$ is the perturbed sample at step $l$. We refer to the pseudocodes in the Algorithms~\ref{pseudocode: faithfulness ti} and ~\ref{pseudocode: faithfulness ts} (Appendix~\ref{appendix: pseudocodes}) for clarification.

It is crucial to be aware of metrics' biases when interpreting the scores of either TS or TI. Faithfulness implementations implicitly favor the saliency method that aligns well with the metric's computation~\cite{covertExplainingRemovingUnified2021}, i.e., TS favors sub-sequences. In contrast, TI avoids this, and practitioners should pick the most suitable one depending on the use case.

\subsection{Sanity}
\label{sec: sanity}

The sanity metric measures the idea that a visual interpretation depends on the interpreted model. Intuitively, suppose the weights and biases of a trained network model were re-initialized with random values. The network's predictions and generated saliency maps should also differ from the original maps. However, this is not always the case. Despite a drop in model accuracy, saliency may remain stable. Hence, we test sanity using a variant of sanity checks proposed by Adebayo et al.~\citep{adebayoSanityChecksSaliency2018}, that performs a \textit{layer-wise cascading randomization} of the network's weights and biases, starting from the output to the input. In contrast to independent randomization, the cascading approach results in a mostly continuous performance degradation of predictions, see Fig.~\ref{fig 1: sane} for an illustration. Network accuracy should increasingly resemble random guessing. Following \cite{adebayoSanityChecksSaliency2018}, we compare saliency using the structural similarity index measure (SSIM)~\citep{wang2004image}, which compares the distribution of sub-sequences of $M^c$. We define the sanity score as 
\begin{equation}\textstyle
    S_{\text{Sanity}}(M, D, m) = -\frac{1}{N} \cdot \sum_{X \in D} \sum_{i=1}^{L} \frac{\text{SSIM}(M_{m}^c(X), M_{m_i}^c(X))}{L},
\end{equation}
where $|D|=N$, $i$ enumerates the $L$ layers of $m$, whose parameters are randomized and $M_{m_i}^c(x)$ is the saliency map after randomizing layer $i$ of $m$. We average the SSIM over $L$ layers and compute the average over all samples in $D$ to reduce the impact of random noise. We refer to the pseudocode in Algorithm~\ref{pseudocode: sanity} (Appendix~\ref{appendix: pseudocodes}) for clarification.

More recently, Yona and Greenfeld~\cite{yonaRevisitingSanityChecks2021} demonstrated that this variant of sanity checks is distribution-dependent and cannot exclusively be relied on when comparing saliency methods. However, the authors do not challenge the sanity check and use it for debugging or explaining models.

\subsection{Inter-Class Sensitivity}
\label{sec: inter-class sensitivity}

In multi-class prediction tasks, the classifier must identify relevant features for each class to make a correct prediction. Hence, the relevance map $M^c$ should identify different salient features for those different classes~\citep{li2021quantitative}, see Fig.~\ref{fig 1: sensible}. If a method is not sensitive to the class, the score may indicate that the classifier failed to learn the correct features for these classes, and the sensitivity score of the saliency should raise a red flag, even though large datasets may contain semantically similar classes. Following ~\cite{rebuffi2020saliency}, we assume that the class with the highest predicted probability is salient, and the one with the lowest confidence is uninformative because it is not contained in the sample. Taking the least confident class thus avoids (semantically) similar samples to be picked, which may well occur in large enough datasets with overlapping categories.
Concretely, inter-class sensitivity~\citep{rebuffi2020saliency} measures class-specific sensitivity of the generated relevance map for the most likely ($c_{\text{max}}$) and least likely ($c_{\text{min}}$) class according to the model. We compute the mean inter-class sensitivity score as
\begin{equation}\textstyle
    S_{\text{Inter-Class Sensitivity}}(M, D) = -\frac{1}{N} \cdot \sum_{X \in D}\operatorname{sim}(M^{c_{\text{max}}}(X), M^{c_{\text{min}}}(X)).
\end{equation}
We compute similarity of two saliency maps as $\operatorname{sim}(M^{c_{\text{max}}}(x), M^{c_{\text{min}}}(x))$ where $\operatorname{sim}(\cdot, \cdot)$ is a similarity function (e.g., a cosine similarity) that is easy to interpret via its geometric interpretation. It is defined as the angle between two non-zero vectors that measures the similarity between their inner product space~\citep{HAN201239_cosine}. Similarity of $M$ in binary classification would result in a negative cosine similarity, meaning nearly inverted saliency maps for max- and min-classes.  We refer to the pseudocode in Algorithm~\ref{pseudocode: inter-class sensitivity} (Appendix~\ref{appendix: pseudocodes}) for clarification.

The sensitivity of the saliency map is not strictly related to how well the saliency method captures model behavior. Saliency methods of low faithfulness could introduce their bias to the sensitivity score and hide actual model behavior. However, inter-class sensitivity is still a helpful indicator of the quality of a model and visualization, with the important caveat that a score is only meaningful if the visualization is both faithful and sane. See Section~\ref{sec: recommendations} for further details and recommendations.

\subsection{Robustness}
\label{sec: robustness}

Saliency methods may be vulnerable to small input changes and even adversarial attacks~\cite{Dombrowski2019,dombrowskiRobustExplanationsDeep2022}. For reliable interpretations, saliency methods should be as robust to small changes in the input as the model itself and represent the model's behavior well.
Hence, we evaluate a method's robustness to noise via its sensitivity of the most likely class ($c_\text{max}$) as proposed by Yeh et al.~\citep{yehFidelitySensitivityExplanations2019}, see Fig.~\ref{fig 1: robust} for an illustration. Intuitively, if a model's predictions are stable even with noisy inputs, the saliency map of a model should not change significantly either~\citep{alvarezmelis2018robustness}, meaning that saliency maps with high sensitivity are less reliable. Yeh et al.~\cite{yehFidelitySensitivityExplanations2019} define the sensitivity of the saliency map derived from the gradient as
\begin{equation}\textstyle
    [\bigtriangledown_{X} M^c(X)]_j = \lim_{\epsilon \rightarrow 0} \frac{M^c(X + \epsilon e_j) - M^c(X)}{\epsilon}
\end{equation}
for any $j \in \{1, ..., |H \cdot T|\}$, where $e_j$$\in$$\mathbb{R}^{H \times T}$ is the $j$-th coordinate vector and the $j$-th entry is one while the others are zero. We use Monte Carlo sampling of $\epsilon e_j$, where $|\epsilon e_j|$$<$$a$ ($a$ is a user-specified radius), to generate different $\hat{X}$ $=$ $X$ $+$ $\epsilon e_j$, effectively reducing the impact of sampling noise. We compare $\hat{X}$ with the original $X$ to compute 
\begin{equation} \label{equ:max_sensitivity}\textstyle
    S_{\text{Max Sensitivity}}(M^c, D, a) = -\frac{1}{N} \cdot \sum_{X \in D} \max_{||\hat{X} - X|| < a} ||M^c(\hat{X}) - M^c(X)||.
\end{equation}
We refer to the pseudocode in Algorithm~\ref{pseudocode: robustness} (Appendix~\ref{appendix: pseudocodes}) for clarification.

Alvarez et al. question whether saliency methods should be robust when the underlying model is not~\citep{alvarezmelis2018robustness}, i.e., if an explanation should include noisy pixels (in image processing). They reason that pixel-exact saliency is helpful for exact debugging, while stable aspects are more valuable for understanding the predictor and the underlying phenomenon. Alternatively, Agarwal et al.~\cite{agarwalRethinkingStabilityAttributionbased2022} proposed relative stability metrics that measure the changes in output explanation w.r.t. a white box model's learned representations.
In summary, especially if the underlying models are not robust, the robustness of saliency methods and models should be jointly evaluated and studied further.

\subsection{Intra-Class Stability}
\label{sec: intra-class stability}
Two explanations of the same feature should be similar, independent of their exact location within the input sample. This is helpful for generalization, as features may be translation-invariant, and predictors should point out the same evidence across samples of the same class~\cite{felHowGoodYour2022}. Fel et al.\cite{felHowGoodYour2022} propose algorithmic stability measures to test whether saliency methods produce similar interpretations for samples from the same class. 

To compute this metric, we need to make several assumptions. First, we assume that the order of temporal features is correlated between different samples of the same class. This is a reasonable assumption since time series data often contains distinct temporal features that follow a dataset-specific order. However, note that scores for intra-class stability are highly dataset-specific and should not be compared across different datasets. See Fig.~\ref{fig: classification datasets} for this evaluation.
Second, we assume that the saliency method is faithful and that both the visualization and the model are robust. This assumption must be validated for every model and visualization technique combination.

Based on these assumptions, we can test the intra-class stability of saliency maps $M^{c}$ for a given dataset $D$ using distance statistics, see Fig.~\ref{fig 1: stable} for an illustration. Specifically, we compute pairwise distances between saliency maps $M^{c}$ for different samples $X_i, X_j \in D$ and aggregate these across classes:
\begin{equation} \label{equ:intra-class-stability}\textstyle
    S_{\text{Intra-Class Stability}}(M^c, D) =-\sum_{i \in [0,N]}\sum_{j \in [i+1,N]}  \frac{d_{\text{dtw}}(M^c(D_i),  M^c(D_j))}{N \cdot(N-1)}.
\end{equation}

We use Dynamic Time Warping (DTW, \cite{vintsyuk1968DTW}) as the distance function $d_{\text{DTW}}$ to account for the time series nature of the data. The score compares each class's sample's $M^c$ with all other samples' saliency maps in the dataset, using $d_{\text{DTW}}$. DTW provides a crucial benefit by comparing time series with similar features in the same causal order but not simultaneously. We refer to the pseudocode in Algorithm~\ref{pseudocode: intra-class stability} (Appendix~\ref{appendix: pseudocodes}) for clarification. In contrast, Fel et al.~\cite{felHowGoodYour2022} evaluate saliency methods' generalization through algorithmic stability. They train two predictor models on different folds, i.e., datasets with and without the sample $X$, and only then compare the predictors' saliency on the $X$.

%
%
%
%
%
%
%
%
%
%
%
%

\section{Experimental Setup}\label{sec: experimental setup}

The experiments first conduct a broad study to develop recommendations for the effective and reliable use of saliency methods on multivariate time series data. Second, it implements the recommendation as part of a study on the tool-use problem.

Our broad study conducts a representative evaluation of saliency methods and saliency metrics for the diverse data that are multivariate time series. Such a broad evaluation is also necessary because saliency methods' explanations may be highly data-dependent. Kindermans et al.~\cite{kindermansReliabilitySaliencyMethods2019} 
showed that explanations may change simply due to applying a global mean shift in data (and features). To achieve a more realistic evaluation, we experiment in various training settings that can change explanations~\cite{dombrowskiRobustExplanationsDeep2022}. We vary the model architectures that may learn different features or their regularization~\cite{aliExplainableArtificialIntelligence2023} that may influence what features the models rely on, see Section~\ref{sec: model architectures}. We also select datasets from diverse domains that can be large, multivariate, and multi-class, and show that saliency methods behave differently depending on the dataset, see Section~\ref{sec: selected datasets}. Finally, we include a broad selection of saliency methods and explain their parameters in Section~\ref{sec: saliency methods}.

\subsection{Model Architectures}\label{sec: model architectures}

We select the two commonly used convolutional model architectures, the Fully Convolutional Network (FCN)~\citep{long2015fully} and the Temporal Convolutional Network (TCN)~\citep{bai2018empirical}. Preliminary studies led us to exclude Multilayer Perceptrons (MLPs) due to noisy saliency and Long Short-Term Memory Networks (LSTMs) ~\citep{HochSchm97} due to vanishing saliency. While transformer-based models can capture global dependencies in long sequences and are becoming more popular for time series classification~\cite{wenTransformersTimeSeries2023}, we exclude them due to two reasons. First, their computational complexity is too high, as the target application involves low-powered integrated devices~\citep{10.1007/978-3-030-67670-4_33_embedded,lofflerAutomatedQualityAssurance2021}. Second, Darcet et al.~\citep{darcet2024vision} have shown that (Vision) Transformer learn feature maps with artifacts in low-informative input areas, that are repurposed for internal computations. This would interfere with saliency map generation, as the artifacts mask input saliency, requiring specialized saliency methods.

Our FCN contains four convolution blocks with a convolutional layer, batch normalization~\citep{bn_Ioffe}, and ReLU activations, similar to~\cite{wang_models_tsc}. The convolutional layers have the kernel shapes \{${7, 5, 3, 3}$\} and numbers of filters \{${16, 32, 32, 16}$\}, with unit stride and no padding. A final fifth block contains a 1x1 convolutional layer without ReLU and serves as a projection layer that reduces the channel size of the feature maps while keeping salient features. Finally, we apply Global Max Pooling~\citep{bai2018empirical} on the feature maps before the softmax operation.

Our TCN consists of four residual blocks with two convolutional layers each. The blocks' convolutions have the kernel shapes \{${7, 5, 3, 3}$\} and numbers of filters \{${16, 32, 32, 16}$\}. We also use Global Max Pooling for the predictions of the TCN~\citep{bai2018empirical}.

We use the Adam optimizer with a learning rate of $0.002$. We train every dataset for $600$ epochs (with early stopping with patience of $80$ epochs) and use a Cross-Entropy loss for classification.

\subsection{Dataset Descriptions}\label{sec: selected datasets}

To develop more general recommendations on the use of saliency methods and metrics on time series data, we have to focus on large, multivariate, multi-class datasets from diverse domains. We validate these guidelines by implementing them on the Tool Tracking dataset. 
We selected the datasets GunPointAgeSpan, FordA, FordB, ElectricDevices, MelbournePedestrian, and NATOPS from the UCR repository~\citep{dau18archive}, based on their diverse set of characteristics and their use in previous studies~\citep{fawaz_tsc, wang_models_tsc, Schlegel2019TowardsAR, ates2020counterfactual}.  While our study focuses on industrial and sensor data, we chose the datasets to be diverse in instance length and time-scale, noise level, missing data and anomalies, periodicity, number of classes, and number of dimensions, imbalance, and periodicity, see Table~\ref{tab:dataset_properties} for details. Due to the application of our recommendations to Tool Tracking, we consider other domains, such as healthcare and finance data, as out of scope.

\begin{table}[h!]
\centering
\caption{Characteristics of the datasets included in the study. $C$ refers to the number of classes, \textit{Size} to the complete data before splitting, \textit{Len.} to a sample's (average) length, \textit{Dim.} to the number of dimensions, \textit{Bal.} to whether a dataset is class-balanced, \textit{Period.} to whether a dataset has periodicity and \textit{Noise} to whether a dataset has medium or higher levels of noise.}
\label{tab:dataset_properties}
\begin{tabular}{|l|c|c|c|c|c|c|c|}
\hline
\textbf{Dataset} & \textbf{C} & \textbf{Size} & \textbf{Len.} & \textbf{Dim.} & \textbf{Bal.} & \textbf{Period.} & \textbf{Noise}  \\ \hline
 FordA & 2 &  4.821 & 500 & 1  & \ding{51}& \ding{55}& \ding{55} \\ \hline
FordB & 2 &  4.446 & 500  & 1  & \ding{51} & \ding{55}& \ding{51}\\ \hline
ElectricDevices & 7 &  16.637 & 96  & 1  & \ding{55} & \ding{51}& \ding{51} \\ \hline
GunPointAgeSpan & 3 & 451 & 150 (\SI{5}{s}) & 1 & \ding{51} & \ding{51} & \ding{55} \\ \hline
MelbournePedestrian & 10 &  3.457 & 24 (\SI{24}{h})  & 1  & \ding{55} & \ding{51} & \ding{55} \\ \hline
NATOPS & 6 &  231 & 47 (\SI{2.3}{s}$\pm 0.6$) & 24  & \ding{51}& \ding{55}& \ding{55} \\ \hline
Tool Tracking & 4 &  3.250 & 31 (\SI{0.2}{s}) & 9  & \ding{55}& \ding{55}& \ding{51} \\ \hline
\end{tabular}
\end{table}

We developed the Tool Tracking dataset~\citep{10.1007/978-3-030-67670-4_33_embedded,lofflerAutomatedQualityAssurance2021} 
to address gaps in the quality assurance process in industrial production lines. Workers performed various assembly tasks with light-weight hand-held tools with attached external sensors. Machine Learning algorithms detect the tool's correct or incorrect usage. To this end, the dataset contains recordings for an electric screw driver.
The sensor package consists of two three-axis inertial (IMU) sensors, i.e., accelerometer and gyroscope, and a three-axis magnetic field sensor. The data contains annotations for classification, where every sample is of one class. We use four classes for the classification task, i.e., tightening and untightening of screws at a work piece, and the classes clockwise and counter clockwise motor activity in the air. Figure \ref{fig: tool tracking} shows examples for the four classes "Tightening", "Untightening", "Clockwise" (CW), and "CounterClockWise" (CCW). Samples correspond to windows of $\SI{0.2}{s}$ duration and are provided with 50\% overlap. The data is multivariate and sampling rates of the sensors are different, with 154.646 Hz for the magnetic field sensor and 102.292 Hz for the IMU, which is up-sampled to match the higher rate. We split the dataset into train and test with a ratio of 75\%/25\% and use 10\% of the training data for validation, see Table~\ref{tab:dataset_sizes} for the dataset size.
\begin{table}[h!]
\centering
\caption{Dataset sizes for the training, validation, and test sets of the Tool Tracking dataset.}
\label{tab:dataset_sizes}
\begin{tabular}{|l|c|c|c|c|c|}
\hline
\textbf{Datasets} & \textbf{Training} & \textbf{Validation} & \textbf{Test} & \textbf{Classes} & \textbf{Dimensions} \\ \hline
Tool Tracking     & 2193                   & 244                      & 813                & 4 & 9                           \\ \hline
\end{tabular}
\end{table}

Finally, we normalize the training data with zero mean shift and a standard deviation of one, and apply the normalization parameters to validation and test data. The classes are imbalanced, with tightening and untightening being the majority classes.

\subsection{Saliency Methods}\label{sec: saliency methods}

We evaluate nine visual interpretation methods that we chose from the three categories of saliency methods. This selection does not limit the results, as the underlying framework and analytical approach are method-agnostic and can be extended to incorporate other techniques as well. We use the implementations provided by Captum~\cite{kokhlikyan2020captum}. We set the hyperparameters for each visual interpretation method according to their recommendations from the literature.

\begin{itemize}
    \item Gradient~\citep{Simonyan2014DeepIC} computes the derivative of the target class score for the input sample to generate the saliency map of the input sample. 
    \item Integrated Gradient~\citep{sundararajan2017axiomatic} uses the linear path method to compute the gradient along the path from a baseline $x'$. We use a zero vector for the baseline. Smilkov et al.~\cite{smilkov2017smoothgrad} suggest the number of steps for the path between 20 and 300. Hence, we use $60$ steps from baseline $x'$ to the original input sample $x$, according to $x = x' + \frac{(x - x')}{N} \cdot n$, $n$ is the current step. 
    \item SmoothGrad~\citep{smilkov2017smoothgrad} computes the input sample's derivative of the target class score and adds the Gaussian noise $\mathcal{N}(0, \sigma^2)$ to the input sample multiple times before computes the gradients from the perturbed samples $(x + \mathcal{N}(0, \sigma^2))$. We set the number of iterations of adding noise to $60$ and the standard deviation of Gaussian noise to $\sigma = 0.2$. 
    \item Guided Backpropagation~\citep{Springenberg2015StrivingFS} guides the backpropagation through layers with ReLU activations by applying a rule that masks gradients of ReLU layers with negative inputs to their ReLU activations during the forward pass. This masks the gradients of inactive neurons, effectively reducing noise in saliency maps.
    \item GradCAM~\citep{8237336} focuses on the feature maps of the last convolutional layer, which are typically smaller than the shape of input samples. Therefore, we use interpolation to up-sample the saliency maps from the last convolution layer to match them to the input samples for visualization in input space. Because of ReLU functions, GradCAM returns only positive relevances.
    \item Guided-GradCAM~\citep{8237336} multiplies the up-sampled saliency maps from GradCAM element-wise with the saliency maps from Guided Backpropagation to compute the final saliency maps.
    \item Layer-Wise Relevance Propagation (LRP)~\citep{10.1371/journal.pone.0130140} uses the baseline $\epsilon$-propagation rule for every model with $\epsilon = 1e-9$. Due to the residual block in the TCN model, the propagated relevance values are summed up ($R = R_1 + R_2$).
    \item LIME~\citep{10.1145/2939672.2939778} uses the cosine distance function as the kernel function with width $w = 5.0$ to weight the perturbed samples and performs $1000$ iterations. We reduce its computational demands by grouping a few temporal neighboring points, i.e., three or four, depending on sample length, as having similar relevance.
    \item Kernel SHAP~\citep{NIPS2017_8a20a862} uses $1000$ iterations, and the number of features along the time dimension is $50$. The sampling of feature perturbation in Kernel SHAP is based on the distribution $p(f) = \frac{(F - 1)}{(f \cdot (F - f))}$, where $f$ is the number of selected features and $F$ is the total number of features in interpretation space.
\end{itemize}

\section{Experimental Results}
\label{sec: experiments}

This section presents the experimental results of the classifier models accuracy in Section~\ref{sec: classification accuracy}, the evaluation of saliency methods with the metrics faithfulness, sensitivity, stability, robustness, sanity in Section~\ref{sec: eval classification}, and a more detailed generalization analysis of the impact of datasets and models on the metrics in Section~\ref{sec: datasets}. The following Section~\ref{sec: recommendations} then develops recommendations and implements them on the tool-use problem. Our study aggregates a total of 672 experiments\footnote{Code available at \url{https://github.com/crispchris/saliency}\label{footnote: github}}.
\subsection{Classification Accuracy}
\label{sec: classification accuracy}

The classification accuracy demonstrates the models' ability to predict datasets, which is fundamental for meaningful saliency maps. 
We present the test accuracy of FCN and TCN on the datasets from the UCR repository, as well as for the Tool Tracking dataset.
The results in Table~\ref{table:train_results} show that FCN and TCN both achieve relatively high scores across most datasets, with ElectricDevices trailing with lower scores. Ideally, a high test accuracy should indicate a more faithful saliency map independent of the saliency method, because the model's learned features are predictive, i.e., generalize well to unseen samples, especially very high accuracy scores of $95.79\pm1.56\%$ for FCN on NATOPS or $95.91\pm2.27$ for TCN on GunPointAgeSpan.
With averages of $84.73\pm1.57\%$ for the FCN and $85.91\pm3.17\%$ for the TCN, and scores in the range between $67.49\pm2.44\%$ and $95.91\pm.2.27\%$, the models demonstrate their ability to predict the selected datasets, and that they should be suitable for realistically scoring saliency methods.

\begin{table}[t]
 \centering
 \begin{tabular}{|rcc|}
    \hline
    \textbf{Dataset} & \textbf{FCN} & \textbf{TCN} \\ 
    \hline
    GunPointAgeSpan & $88.38 \pm 2.44$  & $95.91 \pm 2.27$  \\
    \hline
    FordA & $89.43 \pm 0.31$  & $91.42 \pm 0.35$  \\
    \hline
    FordB & $75.47 \pm 1.71$  & $79.38 \pm 1.31$ \\
    \hline
    MelbournePedestrian & $90.94 \pm 0.98$  & $88.39 \pm 5.23$ \\
    \hline
    NATOPS & $95.79 \pm 1.56$  & $90.00 \pm 7.09$ \\
    \hline
    ElectricDevices & $67.49 \pm 2.44$  & $68.00 \pm 1.96$ \\
    \hline
    \hline
    Tool Tracking & $82.13\pm0.16$ & $82.94\pm1.98$ \\
    \hline
 \end{tabular}
 \caption{Accuracy in percent on test data.}
 \label{table:train_results}
\end{table}

\begin{table}[t]
 \centering
 \begin{tabular}{|r|c|c|c|c|}
    \hline
    \textbf{Class} & \textbf{Precision} & \textbf{Recall} & \textbf{F1-score} & \textbf{Support} \\
    \hline
    \textit{Tightening} & 0.87 & 0.93 & 0.90 & 413 \\
    \textit{Untightening} & 0.90 & 0.84 & 0.87 & 178 \\
    \textit{CW} & 0.77 & 0.63 & 0.70 & 153 \\
    \textit{CCW} & 0.65 & 0.77 & 0.71 & 69 \\
    \hline
    \textbf{Macro avg} & 0.80 & 0.79 & 0.79 & 813 \\
    \hline
 \end{tabular}
 \caption{Classification Report for TCN on imbalanced Tool Tracking dataset for the classes "Tightening", "Untightening", "ClockWise" (CW) and "CounterClockWise" (CCW) shows lower scores for rarer classes.}
 \label{table:classification_report}
\end{table}

\subsection{Evaluation of Saliency Methods}
\label{sec: eval classification}

\begin{figure}[t!]
    \centering
    \begin{subfigure}[b]{\textwidth}
         \includegraphics[width=\textwidth, keepaspectratio]{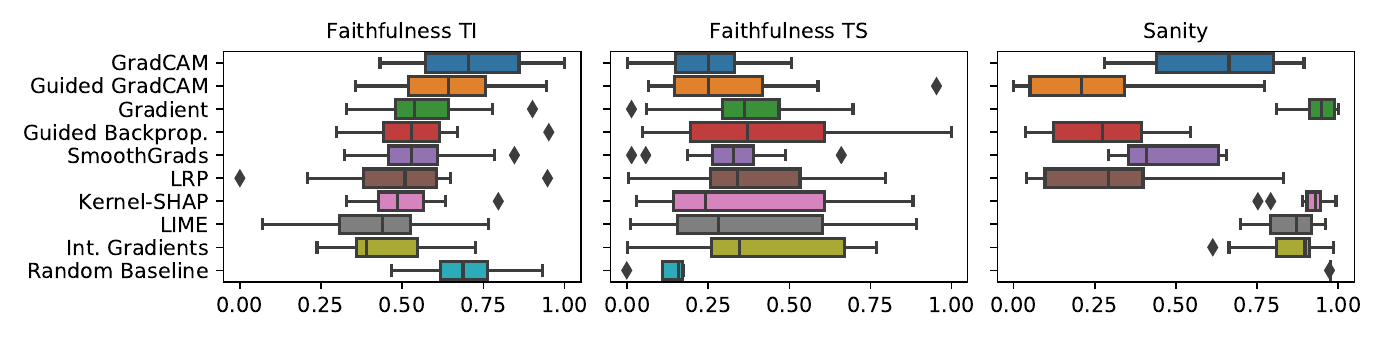} 
    \end{subfigure}
    
    \begin{subfigure}[b]{\textwidth}
         \includegraphics[width=\textwidth, keepaspectratio]{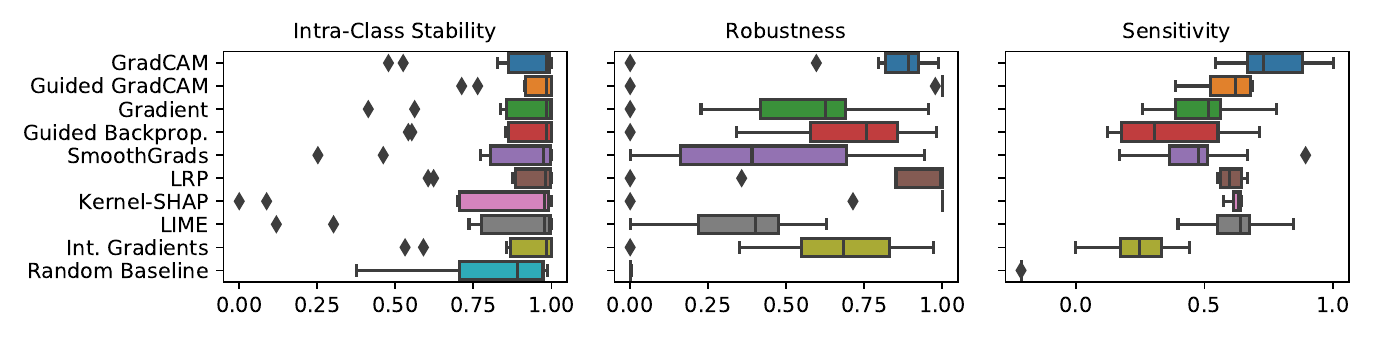}
    \end{subfigure}
    
    \caption{We aggregated results for TCN and FCN overall classification datasets \{model architectures, datasets\} without normalization (as we found that model architectures generally lead to similar scores).}
    \label{fig: classification split}%
\end{figure}

\begin{figure}[t!]
    \centering
    \begin{subfigure}[b]{1.0\textwidth}
         \includegraphics[width=1.0\textwidth,
         keepaspectratio]{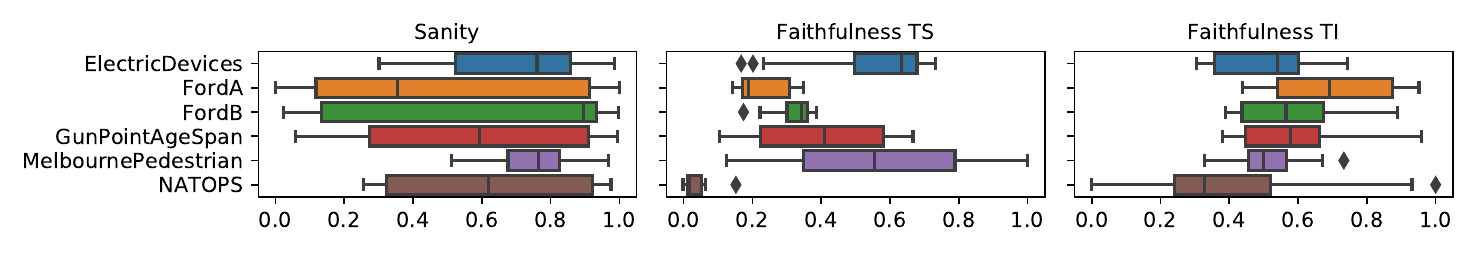}
    \end{subfigure}
    
    \begin{subfigure}[b]{1.0\textwidth}
         \includegraphics[width=1.0\textwidth,
         keepaspectratio]{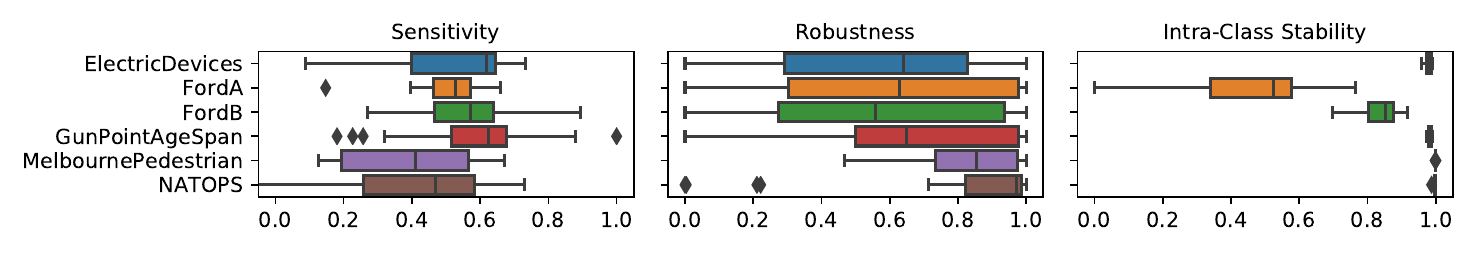}
    \end{subfigure}
    \caption{The plots show the influence of datasets on the scores when we aggregate for each dataset over \{model architectures, methods\}. Similarity metrics for saliency maps yield different results in different domains.}
    \label{fig: classification datasets}%
\end{figure}

This section discusses the evaluation of the saliency methods with our proposed set of quality metrics and forms the basis for the derived recommendations in Section~\ref{sec: recommendations}. To this end, this section first summarizes the scores for each metric over model architectures and datasets in Fig.~\ref{fig: classification split} to highlight the saliency methods' issues. Then, the next Section~\ref{sec: datasets} summarizes the scores for each metric over model architectures and methods in Fig.~\ref{fig: classification datasets} to highlight the dataset-specific issues.

\textbf{Faithfulness.}
We evaluate the saliency metrics under different faithfulness metrics to determine how well the saliency methods capture both individual feature importance and dependencies across sequences.
To test this, we analyze the faithfulness variants Temporal Importance, which assesses the ability to identify individual feature importance, and Temporal Sequence, which evaluates the ability to capture sequential dependencies among features.
Our results reveal significant differences between the two metrics. Saliency methods generally score higher in Temporal Importance, indicating better performance in identifying individual features, but struggle with Temporal Sequence, which measures whether they capture temporal dependencies across sequences. For instance, Fig.~\ref{fig: classification datasets} shows low variance for FordA and NATOPS datasets, suggesting a mismatch with the Temporal Sequence assumption of clustered, sequential features. Especially NATOPS seems to exhibit non-sequentially distributed features. Additionally, there is a notably low correlation between TI and TS. Saliency methods that excelled in TI, specifically GradCAM and Guided GradCAM, only attain relatively modest scores in TS. The more coarse GradCAM even slightly overtakes its Guided variant, likely due to GradCAM capturing the higher-level features and Guided-GradCAM's sample-level interpolation introducing a bias. Integrated Gradients may fail due to the bias introduced by its user-defined baseline.

These findings underscore our hypothesis that the assumptions that Temporal Sequence makes wrt. feature shape may be invalid for certain time series domains. The higher Temporal Importance scores for methods like GradCAM suggest their efficacy in pinpointing individual features, but the lower TS scores highlight a gap in capturing feature dependencies. Overall, the results support the need for visualization methods that can address both individual feature importance and feature sequence dependencies effectively.

\textbf{Sanity.}
The sanity score evaluates the effectiveness of saliency methods in retaining explanations. We test whether the methods produce genuine saliency representing the model's predictive features rather than artifacts after model parameter randomization~\citep{adebayoSanityChecksSaliency2018}.
We compare saliency maps before and after randomizing network parameters.
Our results show that Gradient, Integrated Gradients, LIME, and Kernel SHAP consistently achieve high sanity scores. Conversely, LRP, Guided GradCAM, and Guided Backpropagation maintain similar saliency maps post-randomization, resulting in lower scores.
These findings indicate that certain saliency methods are more reliable in providing meaningful saliency of model predictions. Low sanity methods highlight features in the input that may also be randomly extracted but do not necessarily predict classes, replicating similar results by \cite{adebayoSanityChecksSaliency2018}.

\textbf{Intra-class Stability.}
We test the stability of saliency maps to determine how dataset statistics, such as centered samples as in the GunPointAgeSpan dataset, and inherent feature variance, see Fig.~\ref{results: overview_wodatasets}, affect the consistency of these saliency maps.
We analyze the performance of saliency methods on the selected UCR datasets by assessing their ability to produce stable saliency maps.
Our results reveal that Guided GradCAM consistently generates stable saliency maps across all datasets, likely due to its lower resolution. In contrast, methods like LIME and Kernel SHAP, which consider individual sample points independently, generate less stable saliency maps. Similarly, SmoothGrad, which adds noise to sharpen the saliency maps, can lead to an unintended side effect of over-sharpening and overly focusing the saliency map onto a single, sample-local feature that does not generally represent the rest of the dataset. This, in turn, leads to the low intra-class stability.
High scores for datasets like ElectricDevices and GunPointAgeSpan indicate stable features, while datasets such as FordA exhibit less stability and feature similarity.
In summary, our findings suggest that methods introducing biases, such as Guided GradCAM and Integrated Gradients, tend to generate more stable saliency maps. This stability aligns with datasets having consistent intra-class features. The truthfulness of the saliency, however, should be estimated with a faithfulness metric beforehand. Furthermore, we find that visual analysis of samples can be a helpful indication for the inherent intra-class stability of datasets to estimate dataset quality. Datasets with higher scores exhibit temporal features in specific orders. In contrast, those with low scores generally showed less localized features, indicating issues with the centering of samples within the dataset.
%

\textbf{Robustness.}
To uncover biases introduced by saliency methods, we evaluate the methods with the robustness metric and leave aside a deeper analysis of the underlying model's robustness.
We assess the robustness of the nine saliency methods by introducing random noise into the model’s input and simultaneously recording both the model’s predictions and the saliency map’s sensitivity to these minimal perturbations.
The results show the highest scores for Kernel SHAP, LRP, Guided GradCAM, and GradCAM. Compared to purely gradient-based methods, LRP and the GradCAM variants use class activations, which may have a regularizing effect. Similarly, Kernel SHAP can be considered a more constrained version of LIME. In contrast, the low scores of Gradient confirm findings by Smilkov et al.~\cite{smilkov2017smoothgrad} that show its relatively low robustness. Interestingly, the additive noise of SmoothGrads may compound and lead to diverging saliency maps. Lastly, due to the metric's exceptionally high computational cost, we omit results for \{TCN,ElectricDevices\} that timed out after 12 days of execution.
The results underscore the importance of selecting robust saliency methods that accurately reflect the model's saliency and not introduce method-specific biases.
Our evaluation aims to discover these biases and provide first indications.
However, the metric focuses only on the saliency method's biases due to small changes. Incorporating additional model information, e.g., by scaling saliency appropriately~\citep{agarwalRethinkingStabilityAttributionbased2022}, may ultimately lead to a better understanding of these biases in contrast to any model-specific divergences due to noisy data.

\textbf{Sensitivity.}
We aim to demonstrate that different saliency methods introduce different biases towards class-specific features when applied to different classes within the same dataset.
We compare the inter-class sensitivity of saliency methods across various datasets and model architectures to determine how these methods differ in their response to class-specific features.
The results show that GradCAM exhibits the highest sensitivity to class-specific features, with saliency maps for the most and least likely classes differing significantly. In contrast, Integrated Gradients and Guided Backpropagation yield similar saliency maps regardless of the sample's class. This pattern can be observed consistently across most datasets, as illustrated in Fig.~\ref{fig: classification datasets}.
The results support the conclusion that incorporating biases into the saliency methods has a significant impact on the generated saliency maps. Including class activations, saliency is more sensitive towards class-specific features. In contrast, we observe low sensitivity for Integrated Gradients, which depends on a user-defined baseline. Similarly, Guided Backpropagation intentionally suppresses the negative part of the gradient signal to find more visually focused and "intuitive" interpretations. 
However, as the introduced bias impacts sensitivity, this raises the question whether saliency methods still accurately capture model behavior or instead overly depend on biases.
While these biases can enhance the visual appeal of the saliency maps, our results suggest that they have a non-negligible impact on saliency maps and should be avoided. At the least, saliency methods should first be evaluated for faithfulness and sanity, and sensitivity should be considered secondarily, as a helpful indicator of model behavior.

$~~$

\subsection{Ablation study}
\label{sec: datasets}
We demonstrate whether the choice of saliency methods depends on model architecture and dataset, or whether general recommendations can be made.
We normalize the scores for dataset bias and calculate relative scores between saliency methods to determine their ranking.
The model architecture (FCN or TCN) only contributes a little to the scores, see Fig.~\ref{results: overview_models} in the Appendix. Differences in faithfulness scores may be due to the higher capacity of the TCN, which may learn more diverse features so that its TS score is lower and its TI score is higher. The overall scores' variance and individual outliers in Fig.~\ref{fig: classification datasets} illustrate that datasets heavily impact scores, as the aggregated scores for each dataset over all combinations of \{model architectures, methods\} show. This indicates that the choice of saliency methods depends, for the most part, on the task. 

After normalizing for dataset bias, the relative scores are stable, albeit with noticeable variance, see Fig.~\ref{results: overview_wodatasets}. 
Therefore, we can conclude that the overall choice of saliency method should not typically be influenced by specific architecture of the convolutional or fully convolutional model. 
While we can still confirm the initial ranking of the saliency methods that we presented in Fig.~\ref{fig: classification split}, we closely analysis the datasets' characteristics in the next section and thoroughly discuss how these may impact the choice of saliency methods.

\section{Recommendations}
\label{sec: recommendations}

\begin{figure}[t!] 
    \centering
    \includegraphics[width=0.7\textwidth, keepaspectratio, trim=0 0 0 0.5cm, clip]{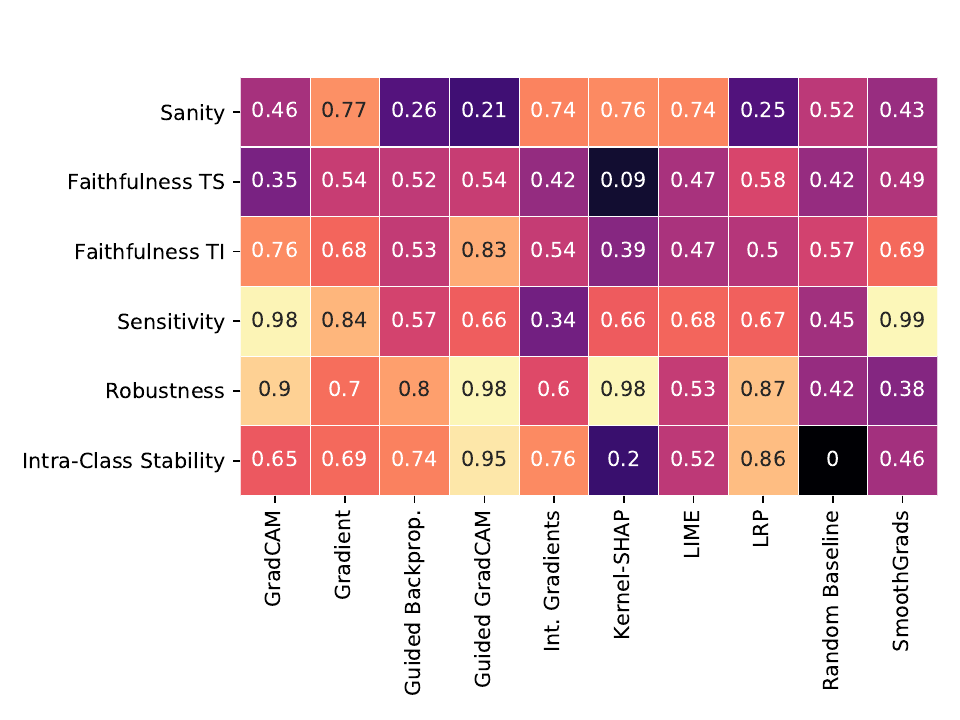}\vspace{-0.2cm}
    \caption{\label{fig: perspective_views} Scores as a heat map for TCN on FordB. Columns with high relative scores (bright squares) indicate good visualizations. Dark squares show the shortcomings.}
\end{figure}

The quality of visualizations differs significantly between methods. No method passes the tests for all categories on all six datasets. This emphasizes the need to evaluate all metrics for every visual interpretation. We recommend using a summary as in Fig.~\ref{fig: perspective_views} to judge visualizations on every category and propose the following insights and guidelines for relative ranking. The absolute scores may be understood compared to a random baseline, similar to shuffled AUC~\citep{Borji_2013_ICCV}.

\subsection{Dataset Characteristics}

We investigate what generalizable insights can be provided, such that practitioners may select an appropriate saliency method based on the nature of their time series dataset. We can make the following observations based on the datasets' characteristics summarized in Table~\ref{tab:dataset_properties}.

\textbf{Imbalanced Datasets}. For datasets like ElectricDevices, MelbournePedestrians, or later in Sec.~\ref{sec: application}, Tool Tracking, it is important to estimate class-wise performance. We report such class-wise F1-scores for Tool Tracking in Table~\ref{table:classification_report}.

The Faithfulness of saliency depends on the model’s accuracy, i.e., higher accuracy implies higher Faithfulness scores. We show the influence of a model's accuracy on the Faithfulness TI metric in Appendix~\ref{appendix: accuracy}, where we varied the regularization parameters to affect the model's performance. This insight is transferable to imbalanced datasets and the more significant F1-scores, e.g., in case that minority classes are harder to predict.

Hence, while our framework proposes aggregate performance indicators for simplicity, we encourage to evaluate saliency methods also class-wise when working with imbalanced datasets, see Sec.~\ref{sec: application} for an example. This helps to better estimate the reliability of saliency for minority classes, such as "CCW" for Tool Tracking.

\textbf{Noisy datasets.} For noisy datasets like ElectricDevices or FordB, the model may have a higher prediction variance. When generalizing to unseen, noisy data, the prediction error is thus larger wrt. variance and has a lower precision. Saliency methods generate a more faithful interpretation for models with higher accuracy, see also Appendix~\ref{appendix: accuracy}. Our results for the Faithfulness TI metric confirm this finding, i.e., models predicting FordA have higher accuracy and Faithfulness TI while those predicting FordB have lower accuracy and Faithfulness TI, see Fig.~\ref{fig: reg plots}. Our conclusion is that the performance of saliency methods is less faithful when applied in the context of (sensor) noise and the visual interpretation of features has to take this into account.


\textbf{Anomalies.} Similarly to noise, anomalies may appear at random in samples. Due to this, an instable Intra-Class Stability metric can hint at such occurrences. As expected, the results for this metric for FordA and FordB show a significantly lower score, see Fig.~\ref{fig: classification datasets}.

\textbf{Multivariate data.} We experiment with the multivariate datasets NATOPS, and later apply our framework to the multivariate dataset Tool Tracking, see Sec.~\ref{sec: application}. NATOPS is challenging for most saliency methods, as the low Faithfulness scores in TS and TI show. We find that only GradCAM achieves a high TI score of between 0.8 and 1.0, depending on model architecture, see Fig.~\ref{appendix fig: classification} (a). This confirms that GradCAM is a suitable choice for multivariate data, such as Tool Tracking, and raises the question on the small number of saliency methods for multivariate datasets. We leave this question for future work.

\textbf{Length and time-scale.} The datasets include lengths of samples in the range from 24 to 500 datapoints per instance and time-scales from milliseconds (Tool Tracking) to hours (MelbournePedestrians). However, there were no findings that we could isolate in our experiments, that show a negative impact of the time-scale, that could not be explained by model performance, dataset noise or multimodality. We conclude that saliency methods are more faithful to model features and less dependent on the time-scale. We conceide that longer sequences, much longer than 500 samples, may produce issues that we do not acount for with our experimental design.


\textbf{Periodicity.} Saliency is especially useful for peridic data, as it helps discovering spurious features, such as sudden spikes, instead of more subtle regular patterns that represent periodicity of the dataset. In our evaluation, we include the datasets ElectricDevices and MelbournePedestrians that include periodic patterns, e.g., electric consumption on a hourly or pedestrian counts on a weekly basis. Hence, we consider our recommendations applicable to such data as well.

While there are more dataset characteristics, such as irregularly sampled data and datasets with missing values, we leave those for future work.

\subsection{Cost Considerations}

The application of a selected saliency method to real-time time series classification may lead to additional cost at inference time, due to each saliency method’s computational complexity. The performance bottlenecks of some methods are less severe (Gradient), whereas others need to be specifically optimized for real-time applications (LIME, KernelSHAP), see Table~\ref{table: computation} for an overview. 

In a real-time scenario, the model prediction and saliency calculation should be shorter than the length of the analyzed data sample. For example, we predict the class of a 200ms sample from the Tool tracking dataset using an FCN on a backend server with a strong NVidia GPU and an AMD 6-core CPU. In this setting, the calculation of the Gradient saliency method takes 1.82 ms and LRP takes \SI{5.05}{ms}. These calculations are well within the time limit, which enables these faster methods to be used in a real-time setting. In contrast, LIME requires 574ms and KernelSHAP needs 549.45ms. 
These cannot be applied in time-critical situations without further optimization of their hyperparameters, such as the reduction of the number of samples of the original model that LIME may use to train its surrogate model. However, such optimization may lead to lower scores in metrics like Faithfulness, limiting their usability. Hence, the saliency methods’ limitations due to their computational complexity should be understood before their practical application.

\subsection{Three-step Guidance}

This section introduces our three-step guidance for ML practitioners on the metrics-guided choice of saliency methods. While these considerations are formulated in a general way, an application may have special dataset characteristics or cost constraints that require close attention to these details.

\textbf{First: Ensure Faithfulness and Sanity.} The general purpose of interpretability methods is to provide insights into model behavior. We propose to use the faithfulness and sanity scores to ascertain that a saliency map represents the model behavior. Faithfulness ensures that the saliency matches the model's predictive features. Note that, depending on the prevalence of temporal correlations and sequential features in a dataset, an expert should choose between TS (when sequential features are essential) and TI (when this is not the case). When in doubt, we propose to use TI.

Sanity checks confirm that saliency maps are sensitive to model parameters. This is important to avoid finding highly salient features, such as edges in images~\citep{adebayoSanityChecksSaliency2018} while being insensitive to model parameters.
We avoid saliency maps with low scores in either metric, as they do not reflect the model's behavior. However, sanity scores have to be seen in the light of discussions on possible effects of gradient shattering that introduces noise into the score~\citep{hedstroem2023sanity}, which lets us prioritize faithfulness (TI) over sanity.
See Fig.~\ref{fig: perspective_views} for an example: Gradient, Integrated Gradients, and LIME achieve the three highest-ranked scores. However, \textit{random saliency maps} (second to the last column) are dissimilar (hence they also show high sanity scores), and their relatively high TI score of 0.57 hints that the faithfulness of the other methods is not entirely reliable, and additional tests are required. Guided Backprop, Guided GradCAM, and LRP fail the sanity check, while Kernel SHAP performs poorly in faithfulness.

\textbf{Second: Check Sensitivity and Robustness.} Once faithfulness and sanity are established, we propose to look at the sensitivity and robustness of the generated saliency maps. A low inter-class sensitivity can indicate that the saliency maps only focus on the predicted class and underestimate the importance of features that do not belong to this class. A low robustness score suggests that the visualization method is susceptible to adversarial examples and small perturbations in the input. Given that the model predictions are robust~\citep{zhang2019trades}, and the saliency is faithful, low robustness implies that the saliency method cannot be trusted and may even be manipulated~\citep{Dombrowski2019}.
We recommend relying less on visual interpretations for non-robust models, especially if they have low faithfulness. According to Fig.~\ref{fig: perspective_views}, we keep Gradient and LIME but disregard Integrated Gradients due to its lower sensitivity.

\textbf{Detailed Analysis: Analyze Intra-class Stability.} Intra-class stability measures how much saliency maps for one class agree between different samples. It allows experts to choose and analyze saliency methods that are more intuitive and understandable, at the cost of trading off their reliability. Note that it is crucial to ensure the faithfulness of a method before relying on this metric. If the method is not faithful, stable visualizations are visually pleasing but do not reflect model behavior. This can hide issues like spurious correlations~\citep{arjovsky2019} or the shortcut learning problem~\citep{geirhos2020shortcut} behind a higher score. 

\textbf{Relative Importance or Weighting.} The most important metric is Faithfulness. It assesses how accurately an explanation reflects the true decision-making process of a model. In light of shattered gradients~\cite{balduzziShatteredGradientsProblem2017}, i.e., the issue of assigning importance of a single parameter’s contribution to predictions in Deep Neural Networks is hard, we recommend to weight Sanity Checks lower relative to the direct measure of a model's Faithfulness.
Furthermore, the secondary metrics, such as Sensitivity or Robustness, require models to be faithful and true to the underlying mechanics. However, there may be a prioritization of metrics of the second and third steps, where a domain expert prioritizes Robustness over Sensitivity in safety-critical applications.

\subsection{Application to Tool-use Time Series}
\label{sec: application}

\begin{figure}[h!] 
    \centering
    \includegraphics[width=1.0\textwidth, keepaspectratio]{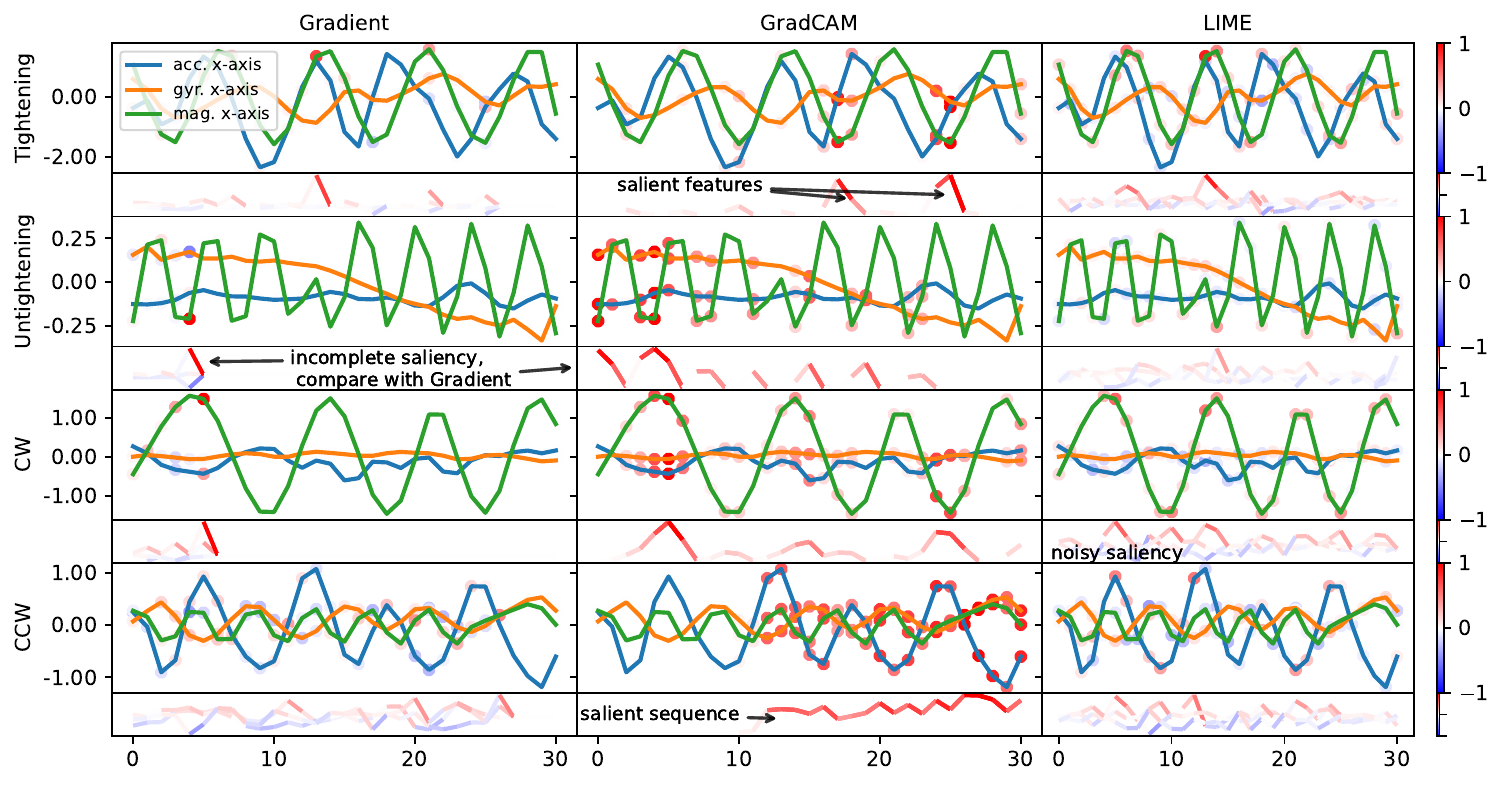}
    \caption{\label{fig: tool tracking} Saliency maps generated by the baseline saliency method Gradient, and GradCAM as the most and LIME as the least faithful (TI). The underlying model is a TCN trained on the Tool Tracking dataset. The figure shows instances of the classes "Tightening", "Untightening", "ClockWise" (CW), and "CounterClockWise" (CCW). Each sample is $\SI{0.2}{s}$ long and contains 31 sensor readings. The sensor data is limited to the x-axis of the sensors accelerometer (acc), gyroscope (gyr), and the magnetic field sensor (mag) for readability, and standardized around the mean zero and a standard deviation of one. Saliency maps are provided in the intervals [-1,1] or [0,1]. The sensor readings are annotated with circles that encode saliency value, and we additionally show the saliency values in the graph below.}
\end{figure}

This section implements our recommendations for the selection of a suitable saliency method for the tool-use problem on the Tool Tracking dataset. We train TCNs and FCNs, calculate saliency maps for the highest-scoring model out of three, and present the scores together with the reasoning for choosing the best saliency method. 
We present one typical sample for each of the four classes in Fig.~\ref{fig: tool tracking} with three select saliency methods on the TCN, illustrating the differences in the maps and scores. We refer to Table~\ref{tab:implementation_results} and Figure~\ref{fig: tool tracking full} in Appendix~\ref{appendix: full results} for additional results.

\textbf{First: Ensure Faithfulness and Sanity.} 
According to our recommendations, we expect Guided GradCAM and GradCAM to lead in faithfulness TI. Our scoring of saliency methods for the TCN confirms the leaf of GradCAM with a score of 0.85 and Guided GradCAM with a score of 0.83, with Gradient behind with 0.65, and finally, LIME trailing with 0.30. These scores indicate the validity of our initial recommendations. Furthermore, they show that CAM-based saliency methods appear more suitable to generate interpretations for the imbalanced, multi-modal Tool Tracking dataset. 

We additionally evaluate the class-wise faithfulness TI score to obtain a more complete understanding of the saliency methods' reliability on the imbalanced Tool Tracking dataset, and report the results in Table~\ref{tab: class-wise ti}. This more detailled analysis is recommended due to the imbalanced nature of the data, e.g., we observe that Gradient's and LIME's saliency maps for the class \textit{Tightening} have very low $S_{\text{Faithfulness TI}}$ of only 0.41 and 0.25 respectively, while GradCAM sores 0.91, indicating its ability to explain the majority class correctly. For brevity, we omit the class-wise evaluation for the remaining saliency metrics.

\begin{table}[ht]
\centering
\begin{tabular}{lccc}
\toprule
\textbf{Metric} & \textbf{Gradients} & \textbf{GradCAM} & \textbf{LIME} \\
\midrule
\textit{Tightening}     & 0.41 & 0.91 & 0.25 \\
\textit{Untightening}     & 0.88 & 0.87 & 0.43 \\
\textit{CW}     & 0.64 & 0.61 & 0.27 \\
\textit{CCW}     & 0.82 & 0.84 & 0.42 \\
\bottomrule
\end{tabular}
\caption{$S_{\text{Faithfulness TI}}$ scores for the three exemplary saliency methods Gradients, GradCAM and LIME. The best overall scores of GradCAM is also reflected in a more reliable class-wise saliency generation. In contrast, Gradients and LIME fail to produce reliable saliency for many or all classes.}
\label{tab: class-wise ti}
\end{table}

Fig.~\ref{fig: tool tracking} illustrates the recommended saliency method GradCAM, that scores the highest faithfulness (TI) score, and contrasts it with the baseline method Gradient and LIME, that has the lowest score. Based on these scores, we can rely on the feature visualization that GradCAM generates with much higher confidence. This qualitative interpretation allows us to explain the possible errors of saliency methods, that our framework helps mitigate. According to LIME, the TCN relies only nearly all points to classify any class. Gradient instead collapses saliency to very few points, such as for the classes "Untightening" and "CW", and to a lesser extend, "Tightening". The most reliable method GradCAM instead highlights salient sequences unique to each class. Predictive features are not focused only on very few or all points in time. Instead, they appear to be change points or value gradients ("Untightening", "CW" "CCW"), minima or maxima ("Tightening", "CW"), or whole sample sequences ("CCW").

Next, we evaluate the sanity score metric and observe the saliency methods Kernel SHAP, GradCAM, LIME, and Integrated Gradients, with average SSIM of below 0.20, while Guided GradCAM has the worst average value of around 0.62. Interestingly, Gradient delivers a worse value compared to the recommendations, which indicates that sanity scores are highly dataset-specific. This confirms the highly variable sanity scores per dataset illustrated in Fig.~\ref{fig: classification datasets}. 
Overall, we consider GradCAM as the most reliable method for a TCN on Tool Tracking.

\textbf{Second: Check Sensitivity and Robustness.} 
Given our recommendations, we expect GradCAM and SmoothGrads to be ahead in Sensitivity if the model learns different salient features for different classes. The largest differences between saliency maps are shown for GradCAM, SmoothGrads, and Guided Backpropagation. However, the semantically similar classes in Tool Tracking may have similar salient features. For that reason, we always have to first rely on faithfulness scores. Hence, select GradCAM and eliminate Guided Backpropagation and SmoothGrads. Interestingly, the high sensitivity score of GradCAM indicates that salient features are class-specific, which is confirmed by visual inspection of the saliency in Fig.~\ref{fig: tool tracking}.

Next, following our recommendations, we expect high scores for GradCAM, Guided GradCAM, and Kernel SHAP on robustness. For the tool-use problem, we observe the strongest robustness for GradCAM, which further confirms its strong performance for this time series data.

\textbf{Detailed Analysis: Analyze Intra-class Stability.} We compare the intra-class stability of saliency generated for a TCN and an FCN to learn about feature stability of the models' learned features. We base the comparison on GradCAM as it was most faithful to both models. We measure distances between the saliency maps of the FCN that are on average 50\% larger than for the TCN. For comparison, intra-class stability for LIME, which has the lowest faithfulness TI score, is nearly identical between TCN and FCN, highlighting the importance of our step-by-step recommendations.

\textbf{Summary.} Overall, our recommendations provide useful step-by-step guidance in selecting GradCAM as the most reliable saliency method for this instance, and allow us to better understand the TCN's learned features as illustrated in Fig.~\ref{fig: tool tracking}. Compared to Gradient and LIME, GradCAM provides more reliable and intuitive insights into the TCN's learned features, such as its focus on large connected segments for the class CCW, or the focus on the min/max values of the magnetic field sensor's readings for the classes Untightening and CW.

\section{Discussion}
\label{sec: discussion}

Most interpretation methods were originally designed for image or text data. Compared to more abstract time series, understanding and comparing visual interpretations is intuitive on image data. Furthermore, the diversity of interpretation methods complicates an objective choice for the model and task~\citep{rojat2021explainable}. For example, when Wang et al.~\cite{wang_models_tsc} and Fawaz et al.~\cite{fawaz_tsc} apply Class Activation Maps (CAM)\citep{Zhou_2016_CVPR_cam} on well-known UCR datasets~\citep{dau18archive}, they notice a qualitative difference of CAM interpretations between network architectures. This suggests that the difference in quality should be quantified. 

This leads to the question of how users should evaluate saliency methods~\citep{li2021quantitative}. Doshi-Velez et al.~\cite{DoshiVelez2017TowardsAR} distinguish saliency metrics as either human-grounded or functional metrics. Human-grounded tests may assess whether an explanation can be quickly understood by a human~\cite{DoshiVelez2017TowardsAR, li2021quantitative}, or ask humans about their opinion on the quality of explanations~\citep{li2021quantitative}, which is costly to scale~\citep{li2021quantitative}. 
Similarly, other work relies on domain experts who perform a qualitative evaluation that is costly compared to an algorithmic one~\citep{Strodthoff_2019_gradinput, Fawaz2019AccurateAI, osti_1613436_interpretable_x_ray}. 

However, human-grounded methods, such as using bounding boxes for the expected location of saliency~\citep{zhang2016EB} can disregard whether saliency methods correctly capture model behavior. They may hide spurious correlations (e.g., "clever Hans" samples~\cite{lapuschkinUnmaskingCleverHans2019, liExperimentalStudyQuantitative2021, andersFindingRemovingClever2022a}). 
Considering this, we can’t help but ask what question the saliency metrics should answer~\citep{DoshiVelez2017TowardsAR, silvaComplementaryExplanationsUsing2018, Carvalho_2019}. Hence, there is a need for objective, quantitative evaluation metrics to make the interpretations' quality measurable and comparable. 

In this line of work, Nguyen et al.~\cite{nguyenRobustExplainerRecommendation2024} recently took an important step in the direction of scoring and ranking saliency methods for time series classification. The authors propose a perturbation-based scoring metric similar to faithfulness (TI). They aggregate scores in their experimental study over different perturbations and classifier models to rank saliency methods. However, relying on only one saliency metric to interpret a model's behavior could be misleading~\citep{tomsettSanityChecksSaliency2020,hedstromQuantusExplainableAI2023}.

In this work, we propose an evaluation scheme for saliency methods based on five orthogonal metrics that provide a disentangled evaluation on time series data, and implement it on the real-world tool-use problem. Our experimental study shows that relying on only one metric, without carefully considering the domain, dataset, and model, may result in interesting but spurious saliency maps. Our recommendations help domain experts understand, rate, and validate saliency for time series in safety-critical applications. 

We find that no individual visualization method achieves high scores on all evaluation methods. For practitioners, we thus recommend focusing on faithfulness (TI) and sanity first.
    Furthermore, the strong impact of datasets on the saliency methods was consistent across all evaluation perspectives and was also evident in the tool-use problem. This indicates that the interpretability of the underlying model is similarly dependent on the learned distribution as its predictive accuracy. 
    Similarly, we found that the saliency methods themselves may introduce an additional bias into the visualization calculations. We show that this can lead to highly stable saliency maps.
    Finally, we found that the model architecture of convolutional networks alone had a relatively low impact on evaluation scores for visualization methods, compared to MLP or LSTM, which were noisy or suffered from vanishing saliency.

Future work investigates the relatively high scores of the random baseline in the two main metrics, sanity and faithfulness. 
Additionally, recent advances point to the importance of latent space evaluation~\cite{schroderWhatLatentSpace2023} and different and potentially improved variants of the faithfulness metric~\cite{turbeEvaluationPosthocInterpretability2023}. 
Our work can be easily extended to other applications, datasets, and model architectures.
If these data modalities and models are vastly different from the ones included in this study, it may be advisable to carefully select appropriate saliency methods~\cite{kokhlikyan2020captum} and subsequently also saliency metrics~\cite{hedstromQuantusExplainableAI2023}.
Finally, model training should be optimized for more reliable interpretability so that it behaves more intuitively without sacrificing predictive performance.

\textbf{Limitations.} However, the primary limitation of this work is the reliance on five orthogonal metrics as the basis for the recommendations. 
First, while these metrics provide a comprehensive framework for evaluating saliency methods, their implementation can be resource-intensive and may pose significant challenges in terms of computational cost and development effort. We address these challenges by making the code available online. 
Second, the development of saliency methods remains highly agile and may produce novel and orthogonal saliency metrics to extend our framework in the future. 
Third, we excluded evaluating the impact of different learning algorithms from our framework and instead relied on the published recommendations. However, the importance of the appropriate selection of optimizers and their parameters can have a significant impact on the models' performance~\cite{iqbalComparativeInvestigationLearning2021}, and following that, the saliency and its scores. Hence future work should also include this dimension.
Fourth, our study focuses on industrial and sensor data. 
We have to limit the scope of our study for practical reasons and select datasets according to our target application of industrial Tool Tracking data. This dataset consists of sensor time series. We chose the datasets we base our recommendations on to be diverse wrt. their characteristics, like instance lengths, time-scales, noise levels, anomalies, periodicity, number of classes, and number of dimensions. However, we acknowledge that this does not cover all possible scenarios, nor does it consider other application domains like healthcare or finance. Still, we believe that future work may successfully transfer of our proposed framework to other application domains.
Finally, since time series data inherently exhibit diverse characteristics, we still recommend that researchers meticulously select saliency methods and metrics for novel applications involving different sensor modalities, particularly those divergent from our own.

\textbf{Impact.} Since the release of the original pre-print version of this manuscript, several studies have engaged with this work~\cite{zhaoInterpretationTimeSeriesDeep2023, fleischmannExploringDatasetBias2025, schroderWhatLatentSpace2023, rezaeiExplanationSpaceNew2025, mengImpletPosthocSubsequence2025}, indicating interest in our evaluation framework. Zhao et al.~\cite{zhaoInterpretationTimeSeriesDeep2023} review saliency methods for time series and  find that framework to be "simple, clear and effective for selecting post-hoc explanations". Furthermore, Rezaei et al.~\cite{rezaeiExplanationSpaceNew2025} acknowledge that we show that "no single saliency method can outperform others in all metrics and across all datasets". Our work was applied in healthcare, i.e., for gait biomechanics~\cite{fleischmannExploringDatasetBias2025}, and influenced subsequent research~\cite{schroderWhatLatentSpace2023,rezaeiExplanationSpaceNew2025,mengImpletPosthocSubsequence2025}.





%
%
%
%
%
%
%
%
%
%
%
%
\section{Conclusion}
\label{sec: conclusion}

This work proposes an evaluation scheme that contributes a rigorous assessment of saliency methods for multivariate time series classification and implements it on the tool-use time series problem.
We recommend using five orthogonal metrics instead of relying on only one aggregate value. Our proposed set of metrics calculates disentangled scores to rate saliency methods. We provide an extensive empirical evaluation that includes nine gradient-, propagation,- or perturbation-based post-hoc saliency methods, six time series datasets from the UCR repository, two deep convolutional network architectures, and an implementation on the multivariate Tool Tracking dataset.
This paper demonstrates how we distinguish between plausible but wrong saliency maps using the provided recommendations and following their step-by-step guidance. The resulting saliency maps are intuitive because they are faithful, are produced by sane saliency methods, have meaningful sensitivity to class-specific features, are robust (if the model is robust), and are stable (if the model operation is consistent). With our contributions, we strive to encourage researchers to rigorously validate visual interpretations of time series models and rely on objective and reliable explanations.

\backmatter

\section*{Declarations}

\bmhead{Funding} Partial financial support was received from Pontificia Universidad Católica de Valparaíso, Vice Rector's Office for Research, Creation and Innovation within the programs "DI Iniciación" (039.485/2024) and "Semilla 2023" (039.247), by the IFI program of the German Academic Exchange Service (DAAD), the Bavarian Ministry of Economic Affairs, Infrastructure, Energy and Technology as part of the Bavarian project Leistungszentrum Elektroniksysteme (LZE) and through the Center for Analytics-Data-Applications (ADA-Center) within the framework of "BAYERN DIGITAL II".
\bmhead{Competing interests} The authors declare that they have no competing interests
\bmhead{Ethics approval and consent to participate} Not applicable
\bmhead{Consent for publication} Yes
\bmhead{Data availability} Published on GitHub under \url{https://github.com/mutschcr/tool-tracking}
\bmhead{Materials availability} Not applicable
\bmhead{Code availability} Published on GitHub under \url{https://github.com/crispchris/saliency}
\bmhead{Author contribution} Christoffer Löffler, Wei-Cheng Lai, Dario Zanca, and Christopher Mutschler contributed to the study conception and design. Material preparation and data collection were performed by Christoffer Löffler. The machine learning methodology was contributed by Christoffer Löffler and Wei-Cheng Lai. The first draft of the manuscript was written by Christoffer Löffler and Wei-Cheng Lai, and all authors extended and commented on previous versions of the manuscript. All authors read and approved the final manuscript.

\begin{appendices}




\section{Pseudocodes}
\label{appendix: pseudocodes}

We describe each saliency metric with pseudocode to clarify the algorithms' implementations. Please find the link to our released source code in footnote~\ref{footnote: github}. In this section, we outline the metric Faithfulness TI in Algorithm~\ref{pseudocode: faithfulness ti}, Faithfulness TS in Algorithm~\ref{pseudocode: faithfulness ts}, Sanity Check in Algorithm~\ref{pseudocode: sanity}, Inter-Class Sensitivity in Algorithm~\ref{pseudocode: inter-class sensitivity}, Robustness in Algorithm~\ref{pseudocode: robustness}, and Intra-Class Stability in Algorithm~\ref{pseudocode: intra-class stability}.

\begin{algorithm}
\caption{Faithfulness Evaluation using Temporal Importance (TI)}
\begin{algorithmic}[1]
\Require Model $M$, Dataset $D$, Target class $c$, Data length $L$
\State Initialize $S_{\text{Faithfulness TI}} \gets 0$
\State Compute $N \gets |D|$ \Comment{Number of samples in the dataset}
\State Compute $\text{mean} \gets \text{ComputeMean}(D)$ \Comment{Compute mean values}
\For{each sample $X \in D$}
    \State Identify $(t, i)$ with max relevance in $X$ \Comment{Indices $t$ for time points and $i$ for features}
    \For{$l \gets 0$ to $L-1$} 
        \State Perturb $X'$ by replacing $X[t][i]$ with $\text{mean}[t]$ \Comment{According to relevances' $(t, i)$ in desc. order, until $X'$ is fully perturbed}
        \State $m^c(X'_l) \gets predict(M, X'_l, c)$ \Comment{Calculate softmax prediction}
        \State Update $S_{\text{Faithfulness TI}} \gets S_{\text{Faithfulness TI}} - \frac{1}{N \cdot L}m^c(X'_l)$
    \EndFor
    
\EndFor

\Return $S_{\text{Faithfulness TI}}$
\end{algorithmic}
\label{pseudocode: faithfulness ti}
\end{algorithm}

\begin{algorithm}
\caption{Faithfulness Evaluation using Temporal Sub-sequences (TS)}
\begin{algorithmic}[1]
\Require Model $M$, Dataset $D$, Target class $c$, Perturbation length $L$
\State Initialize $S_{\text{Faithfulness TS}} \gets 0$
\State Compute $N \gets |D|$ \Comment{Number of samples in the dataset}
\State Compute $\text{mean} \gets \text{ComputeMean}(D)$ \Comment{Compute mean values}
\For{each sample $X \in D$}
    \State Identify $(t, i)$ with max relevance in $X$ \Comment{Indices $t$ for time points and $i$ for features}
    \State Perturb $X'$ by replacing sub-sequence around $t$ of length $L$ with mean values
    \State $\Delta m^c \gets m^c(X) - m^c(X')$ \Comment{Calculate softmax predictions' difference}
    \State Update $S_{\text{Faithfulness TS}} \gets S_{\text{Faithfulness TS}} + \frac{1}{N}\Delta m^c$
\EndFor

\Return $S_{\text{Faithfulness TS}}$
\end{algorithmic}
\label{pseudocode: faithfulness ts}
\end{algorithm}

\begin{algorithm}
\caption{Sanity Metric Evaluation using Layer-wise Cascading Randomization}
\begin{algorithmic}[1]
\Require Model $M$, Dataset $D$, Saliency function $m$, Number of layers $L$
\State Initialize $S_{\text{Sanity}} \gets 0$
\State Compute $N \gets |D|$ \Comment{Number of samples in the dataset}
\For{each sample $X \in D$}
    \State $\text{SSIM}_{\text{sum}} \gets 0$
    \For{$i \gets 1$ to $L$} \Comment{Enumerate the $L$ layers of $m$}
        \State $M' \gets random(M,i)$ \Comment{Cascading randomization of parameters up to and including layer $i$ in model $M$}
        \State $S_{\text{orig}} \gets M_m^c(X)$ \Comment{Compute original saliency map}
        \State $S_{\text{rand}} \gets M_{m}^{'c}(X)$ \Comment{Compute saliency map after randomization}
        \State $\text{SSIM}_{\text{sum}}  \gets \text{SSIM}_{\text{sum}}  + \frac{\text{SSIM}(S_{\text{orig}}, S_{\text{rand}})}{L}$
    \EndFor
    \State $S_{\text{Sanity}} \gets S_{\text{Sanity}} - \frac{\text{SSIM}_{\text{sum}} }{N}$ \Comment{Average over all samples}
\EndFor

\Return $S_{\text{Sanity}}$
\end{algorithmic}
\label{pseudocode: sanity}
\end{algorithm}

\begin{algorithm}
\caption{Inter-Class Sensitivity Metric Evaluation}
\begin{algorithmic}[1]
\Require Model $M$, Dataset $D$, Saliency function $m$, Similarity function $\operatorname{sim}(\cdot, \cdot)$
\State Initialize $S_{\text{Inter-Class Sensitivity}} \gets 0$
\State Compute $N \gets |D|$ \Comment{Number of samples in the dataset}
\For{each sample $X \in D$}
    \State $P \gets M(X)$ \Comment{Compute predicted probabilities for each class}
    \State $c_{\text{max}} \gets \arg\max_{c} P_c$ \Comment{Identify most likely class}
    \State $c_{\text{min}} \gets \arg\min_{c} P_c$ \Comment{Identify least likely class}
    \State $S_{c_{\text{max}}} \gets M_m^{c_{\text{max}}}(X)$ \Comment{Compute saliency map for $c_{\text{max}}$}
    \State $S_{c_{\text{min}}} \gets M_m^{c_{\text{min}}}(X)$ \Comment{Compute saliency map for $c_{\text{min}}$}
    \State $\text{similarity} \gets \operatorname{sim}(S_{c_{\text{max}}}, S_{c_{\text{min}}})$ \Comment{Compute similarity}
    \State Update $S_{\text{Inter-Class Sensitivity}} \gets S_{\text{Inter-Class Sensitivity}} - \frac{\text{similarity}}{N}$ \Comment{Average over all samples}
\EndFor

\Return $S_{\text{Inter-Class Sensitivity}}$
\end{algorithmic}
\label{pseudocode: inter-class sensitivity}
\end{algorithm}

\begin{algorithm}
\caption{Robustness: Max Sensitivity Metric Evaluation}
\begin{algorithmic}[1]
\Require Model $M$, Dataset $D$, Saliency function $m$, Radius $a$
\State Initialize $S_{\text{Max Sensitivity}} \gets 0$
\State Compute $N \gets |D|$ \Comment{Number of samples in the dataset}
\For{each sample $X \in D$}
    \State $P \gets M(X)$ \Comment{Compute predicted probabilities for each class}
    \State $c_{\text{max}} \gets \arg\max_{c} P_c$ \Comment{Identify most likely class}
    \State Initialize $\text{max\_difference} \gets 0$
    \State $S_{orig} \gets M_m^{c_{\text{max}}}(X)$ \Comment{Saliency map of original $X$}
    \For{each $j \in \{1, \ldots, |H \cdot T|\}$} \Comment{Generate different perturbations}
        \State $ e_j \gets M[j]$ \Comment{$e_j$ is the $j$-th coord. vector, $j$-th entry is one, others zero}
        \State $\epsilon \gets \text{MonteCarlo}(e_j, a)$ with $|\epsilon e_j| < a$ \Comment{Monte-Carlo Sampling}
        \State $\hat{X} \gets X + \epsilon e_j$ \Comment{Generate perturbed input}
        \State $S_{perturbed} \gets M_m^{c_{\text{max}}}(\hat{X})$ \Comment{Saliency map of perturbed $\hat{X}$}
        \State $\text{difference} \gets \left\lVert S_{orig} - S_{perturbed} \right\rVert$ \Comment{Compute saliency difference}
        \If{$\text{difference} > \text{max\_difference}$}
            \State $\text{max\_difference} \gets \text{difference}$ \Comment{Store max difference of Saliency Maps}
        \EndIf
    \EndFor
    \State $S_{\text{Max Sensitivity}} \gets S_{\text{Max Sensitivity}} - \frac{\text{max\_difference}}{N}$ \Comment{Average over all samples}
\EndFor
\Return $S_{\text{Max Sensitivity}}$
\end{algorithmic}
\label{pseudocode: robustness}
\end{algorithm}

\begin{algorithm}
\caption{Intra-Class Stability Metric Evaluation}
\begin{algorithmic}[1]
\Require Model $M$, Dataset $D$, Saliency function $m$, Target class $c$, Distance function $d_{\text{DTW}}$ (Dynamic Time Warping)
\State Initialize $S_{\text{Stability}} \gets 0$
\State Compute $N \gets |D|$ \Comment{Number of samples in the dataset}
\For{each class $c$ in dataset $D$}
    \State $D_c \gets \{X \in D \mid \text{label}(X) = c\}$
    \State $N_c \gets |D_c|$ \Comment{Number of samples in class $c$}
    \For{each pair $(X_i, X_j)$ in $D_c$ with $i < j$}
        \State $S_i \gets M_m^c(X_i)$ \Comment{Compute saliency map for $X_i$}
        \State $S_j \gets M_m^c(X_j)$ \Comment{Compute saliency map for $X_j$}
        \State $d \gets d_{\text{DTW}}(S_i, S_j)$ \Comment{Compute distance}
        \State Update $S_{\text{Stability}} \gets S_{\text{Stability}} - \frac{d}{N_c \cdot (N_c - 1)}$
    \EndFor
\EndFor
\Return $S_{\text{Stability}}$
\end{algorithmic}
\label{pseudocode: intra-class stability}
\end{algorithm}

\section{Additional Results for Saliency Methods}
\label{appendix: computation}

Table~\ref{table: computation} shows the execution time of the different saliency methods for the FCN model on the Tool Tracking dataset over a repetition of 218 runs. Each sample of the dataset is about 20ms long. The range of the runtimes is between about \SI{1}{ms} and \SI{574}{ms}, which indicates that only some methods can be used in a real-time application, because the prediction as well as the saliency method require time from the system's performance envelope.

\begin{table}[b]
\centering
\caption{Measured execution times [ms] for saliency methods over 218 runs (FCN model on Tool Tracking).}
\label{table: computation}
\begin{tabular}{|l|r|}
    \hline
    \textbf{Method} & \textbf{ms} \\ \hline
    GradCAM & 1.11 \\ \hline
    Guided Backprop & 1.58 \\ \hline
    Gradient & 1.82 \\ \hline
    Smooth Gradients & 1.83 \\ \hline
    Guided GradCAM & 2.18 \\ \hline
    Integrated Gradient & 2.53 \\ \hline
    LRP & 5.051 \\ \hline
    KernelShap & 549.45 \\ \hline
    LIME & 574.71 \\ \hline
\end{tabular}
\end{table}

\section{Additional Results for Framework}

\begin{sidewaysfigure}[ht]
    \centering
        \begin{subfigure}[b]{01.0\textwidth}
         \centering
         \includegraphics[width=1.\textwidth,keepaspectratio]{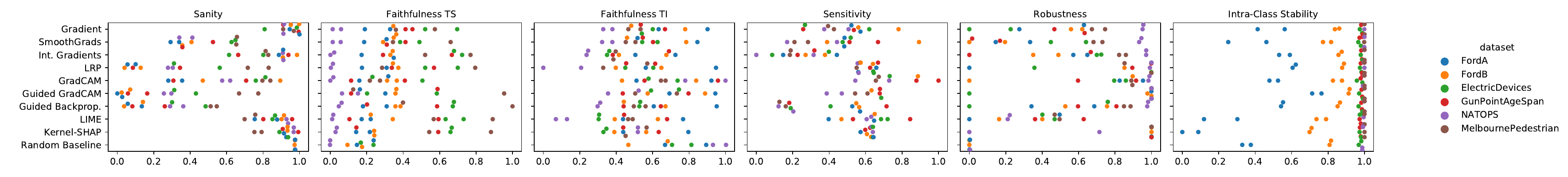}
         \caption{Separated datasets.}
         \label{results: swarmplot}
    \end{subfigure}

    \begin{subfigure}[b]{1.0\textwidth}
         \includegraphics[width=1.\textwidth, keepaspectratio]{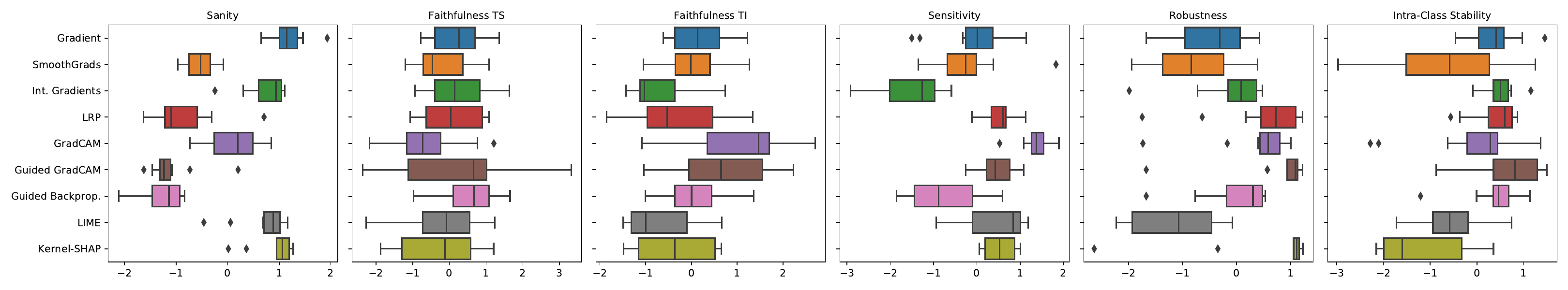}
         \caption{Relative scores that a method can achieve compared to other methods on the same datasets.}
         \label{results: overview_wodatasets}
    \end{subfigure}
    
    \begin{subfigure}[b]{1.0\textwidth}
         \includegraphics[width=1.\textwidth, keepaspectratio]{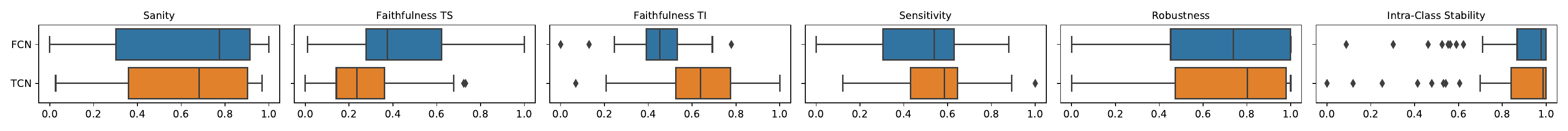}
         \caption{Results for different model architectures, aggregated across datasets and visual interpretability methods. Model architecture has a comparatively small influence on the performance, which speaks to their generalization capabilities.}
         \label{results: overview_models}
    \end{subfigure}
    
    \caption{
     (\subref{results: swarmplot}) shows scores separately over all \{model, method \} combinations. 
     (\subref{results: overview_wodatasets}) removes the datasets' bias to assess the performance of individual visual interpretability metrics independent of datasets.
     (\subref{results: overview_models}) shows results for different model architectures aggregated across datasets and visual interpretability methods. Model architecture has a comparatively small influence on the performance, which speaks to their generalization capabilities. 
    }
    \label{appendix fig: classification}%
\end{sidewaysfigure}

We report additional results to supplement our discussion for the classification task. We show the variability of each visualization method for each dataset and metric category for the TCN in Fig.~\ref{results: swarmplot}. Notably, some datasets are more problematic for the methods, especially for faithfulness and intra-class stability, than others. We also show that the model has only a small influence. For this result, we aggregate all datasets and visualization methods separately for FCN and TCN in Fig.~\ref{results: overview_models}. 

To compare methods independent of datasets, we control for their bias by normalizing the scores of all methods for each \{category, dataset\} so that their mean is $0$ and variance is $1$. With this, we can correctly assess the "relative" performance of the visual interpretation methods across different datasets. The plot in Fig.~\ref{results: overview_wodatasets} shows these relative (or marginal) scores that a method can achieve compared to other methods on the same datasets.

We prove the orthogonality of metrics in Figure~\ref{appendix fig: correlations}. No combination of our metric scores shows a high correlation. This indicates that each metric measures a quality independent of all other metrics.

We show examples of the saliency of the TCN architecture on the FordB dataset for the visual interpretation methods for class 0 in Figure~\ref{appendix fig: example class 0} and class 1 in Figure~\ref{appendix fig: example class 1}. We argue that our framework should guide the choice of a suitable method to generate saliency maps.

\section{Additional Results on Accuracy}
\label{appendix: accuracy}

\begin{figure}[h!]
    \centering
    \includegraphics[width=1.0\textwidth]{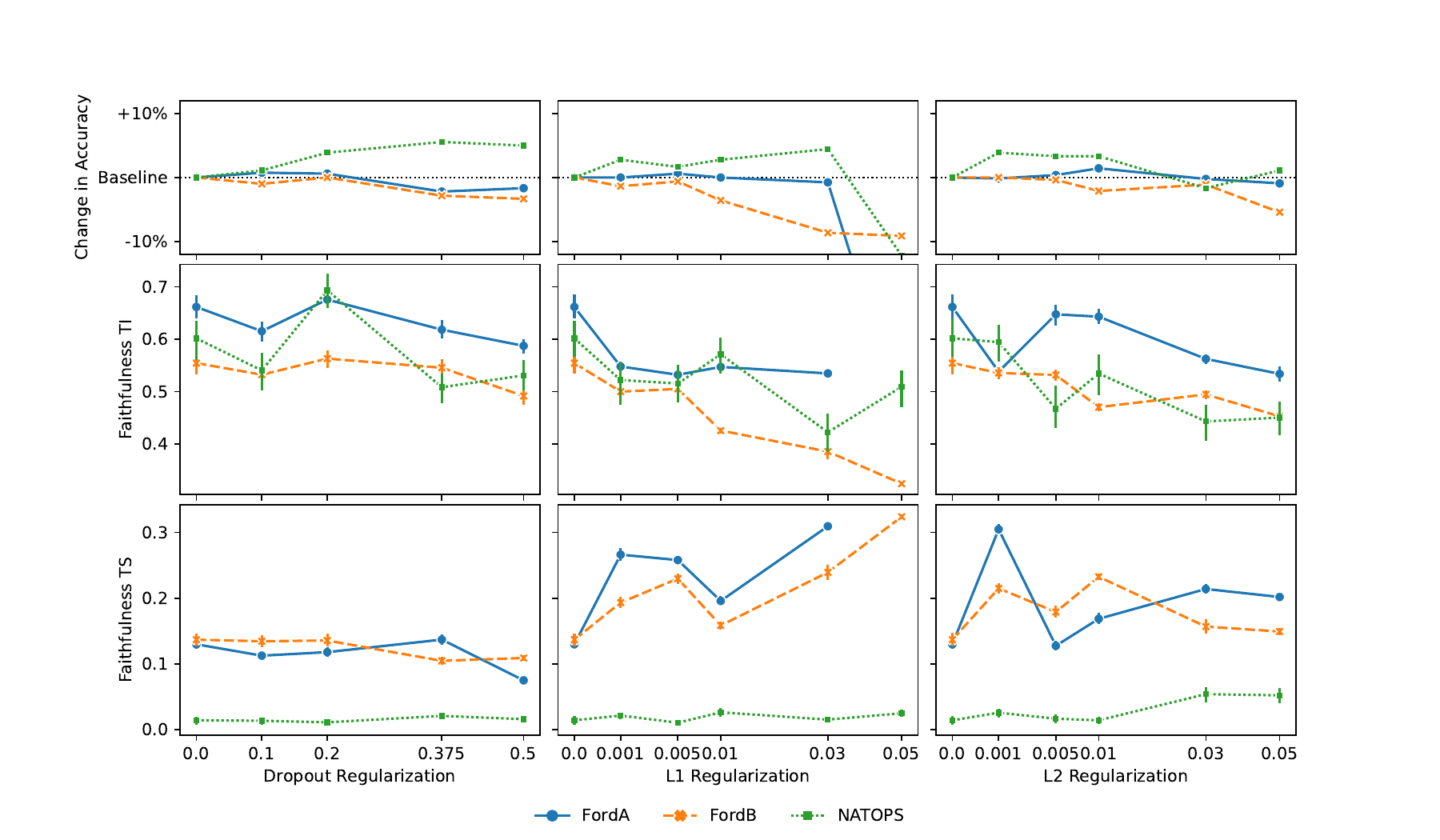}
    \caption{
    Accuracy (top), faithfulness TI (middle), and TS (bottom) scores for all visualizations across the FordA (blue), FordB (orange), and NATOPS (green) datasets. 
    }
    \label{fig: reg plots}%
    
\end{figure}

\textbf{Hypotheses.} We hypothesize differences in faithfulness due to regularization effects. \textit{Dropout regularization} can be understood as computing the average of several thinned networks~\citep{10.5555/2627435.2670313_dropout}. Regularized models rely less on individual features. Hence, models should score higher in Temporal Importance because of TI's arbitrary choice of input features of the highest saliency and lower for Temporal Sequence, which is biased towards clusters of important features. Weight decay~\citep{plautWeigh1986}, such as $\ell_1$ or $\ell_2$ regularization, should result in the opposite learning effect. An $\ell_1$ loss term causes sparsity of the parameters and supports finding the smallest required feature set, which may also influence saliency~\cite{aliExplainableArtificialIntelligence2023}.
Models may learn fewer features that are of higher importance~\citep{krogh1991simple}. This should result in higher scores for TS. More specifically, $\ell_2$ regularization optimizes models to learn more features of similar importance, thus equalizing the importance of learned features~\citep{plautWeigh1986}. Hence, their saliency's faithfulness should perform similarly to the TI and TS variants.

\textbf{Accuracy.} Fig.~\ref{fig: reg plots} shows how regularization impacts the accuracy of an FCN on FordA, FordB, and NATOPS in the top row of plots. We show the relative change from the non-regularized baseline, indicated by the hyper-parameter $\lambda=0.0$, and average the results over two random seeds. Differences in accuracy vary between dropout, $\ell_1$, and $\ell_2$ and also between datasets.
The accuracy on the FordA dataset is relatively stable with regularization, except for an outlier at the large $\ell_1$ reg. of $\lambda=0.05$. In contrast, the accuracy declines on FordB, with a downward trend for increasing values of $\lambda$. Interestingly, the FCN benefits from regularization on the more complex NATOPS dataset. Dropout reliably improves accuracy, while the positive effect of weight decay is not as stable for larger $\lambda$ of $0.05$ ($\ell_1$) and $0.03$ ($\ell_2$).
The effect of a too large $\ell_1$ weight decay is most detrimental; the model even fails to learn the two-class task FordA.

\textbf{Absolute Faithfulness.} We show the saliency methods' aggregated Faithfulness TI and TS scores. Despite similar accuracy for the regularization schemes, faithfulness scores differ strongly. For TI, a small dropout regularization produces better saliency than $\ell_1$ or $\ell_2$ regularization across all datasets, even though the accuracy for models is close, e.g., models for FordA perform within \SI{1}{\%}. Interestingly, the opposite holds for TS, where $\ell_2$ regularization positively affects the FordA and FordB datasets. $\ell_1$ regularization displays a similar effect on all datasets. This confirms our hypotheses that enforcing feature sparsity can affect faithfulness TI and TS differently.
We observe outliers for dropout regularization with $\lambda=0.2$, which markedly improved faithfulness scores across datasets.
A meaningful faithfulness score could not be computed for $\ell_1$ regularization with $\lambda=0.05$ on FordA because the model did not exceed the performance of random guessing.

\textbf{Accuracy and Faithfulness.} Interestingly, we do not see that an increase in accuracy, e.g., for NATOPS or $\ell_2$ with $\lambda=0.01$, consistently correlates with an increase in a TI score. This may be due to the limited fraction of input values (we use 20\%) that we perturb while calculating TI.
However, a decrease in accuracy tends to come with worse TI. Additionally, to less accurate models, there may be a second effect at play. The thinning of features due to weight decay leads to fewer learned features. Perturbing less predictive features naturally results in lower TI scores. The FCN's concentration on fewer features with $\ell_1$ regularization is especially apparent in the Ford datasets' trends in TS. This faithfulness variant is biased towards clustered features and benefits from feature sparsity.

We conclude from these experiments that regularization is vital for accuracy and faithful saliency. Any faithfulness directly depends on the relationship between the accuracy of the models and saliency maps, even if it is not instantly apparent when using a fixed fraction of perturbation when calculating TI because regularization may increase or reduce feature sparsity in the model. Practitioners should choose a suitable regularization depending on the faithfulness variant. Similar test accuracy does not automatically imply similar faithfulness of saliency. This may be due to the numerical effects of weight decay and dropout on the networks' parameters and feature learning.

\section{Additional Results on Tool Tracking}
\label{appendix: full results}
We show the full results for the implementation of our recommendations on the Tool Tracking dataset in this appendix. These additional scores and saliency maps support the discussion in Section~\ref{sec: application}.

In Table~\ref{tab:implementation_results}, we show the saliency scores of the five metrics (faithfulness (TI), sanity, sensitivity, robustness and stability) for the nine saliency methods. We did not normalize the scores as in Section~\ref{sec: eval classification} but instead show each value as calculated for the TCN on the Tool Tracking dataset to facilitate its interpretation.

\begin{sidewaysfigure}
    \begin{figure}[H] 
        \centering
        \includegraphics[width=1.0\textwidth, keepaspectratio]{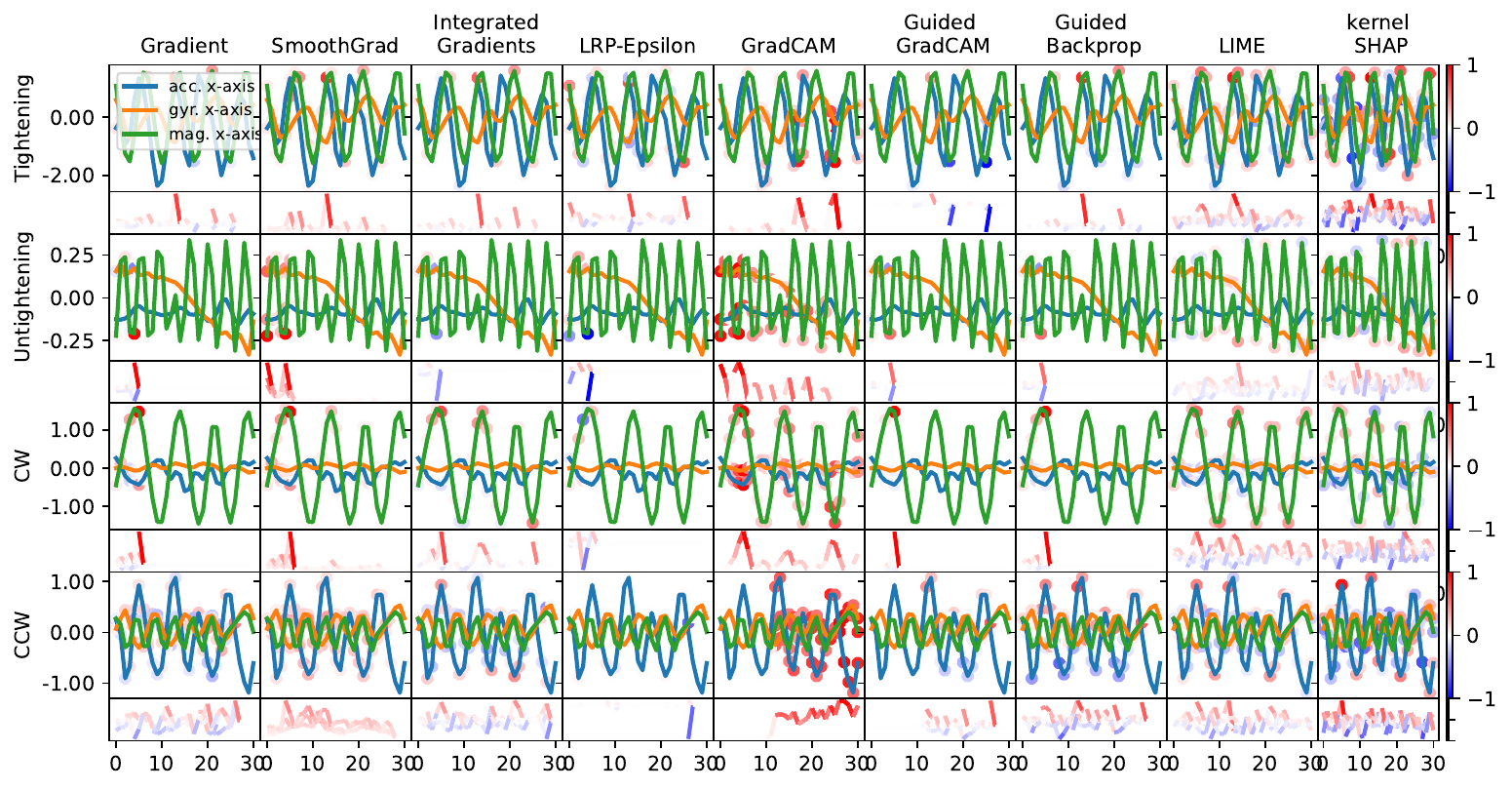}
        \caption{\label{fig: tool tracking full} Saliency maps generated by nine saliency methods. The underlying model is a TCN trained on the Tool Tracking dataset. The figure shows instances of the classes "Tightening", "Untightening", "ClockWise" (CW) and "CounterClockWise" (CCW). Each sample is $\SI{0.2}{s}$ long and contains 31 sensor readings. The sensor data is limited to the x-axis of the sensors accelerometer (acc), gyroscope (gyr) and the magnetic field sensor (mag) for readability, and standardized around the mean zero and a standard deviation of one. Saliency maps are provided in the intervals [-1,1] or [0,1], and the sensor reading additionally annotated with circles that encode saliency value.}
    \end{figure} 
\end{sidewaysfigure}

\begin{figure}[ht]
    \centering
     \includegraphics[width=1.0\textwidth, keepaspectratio]{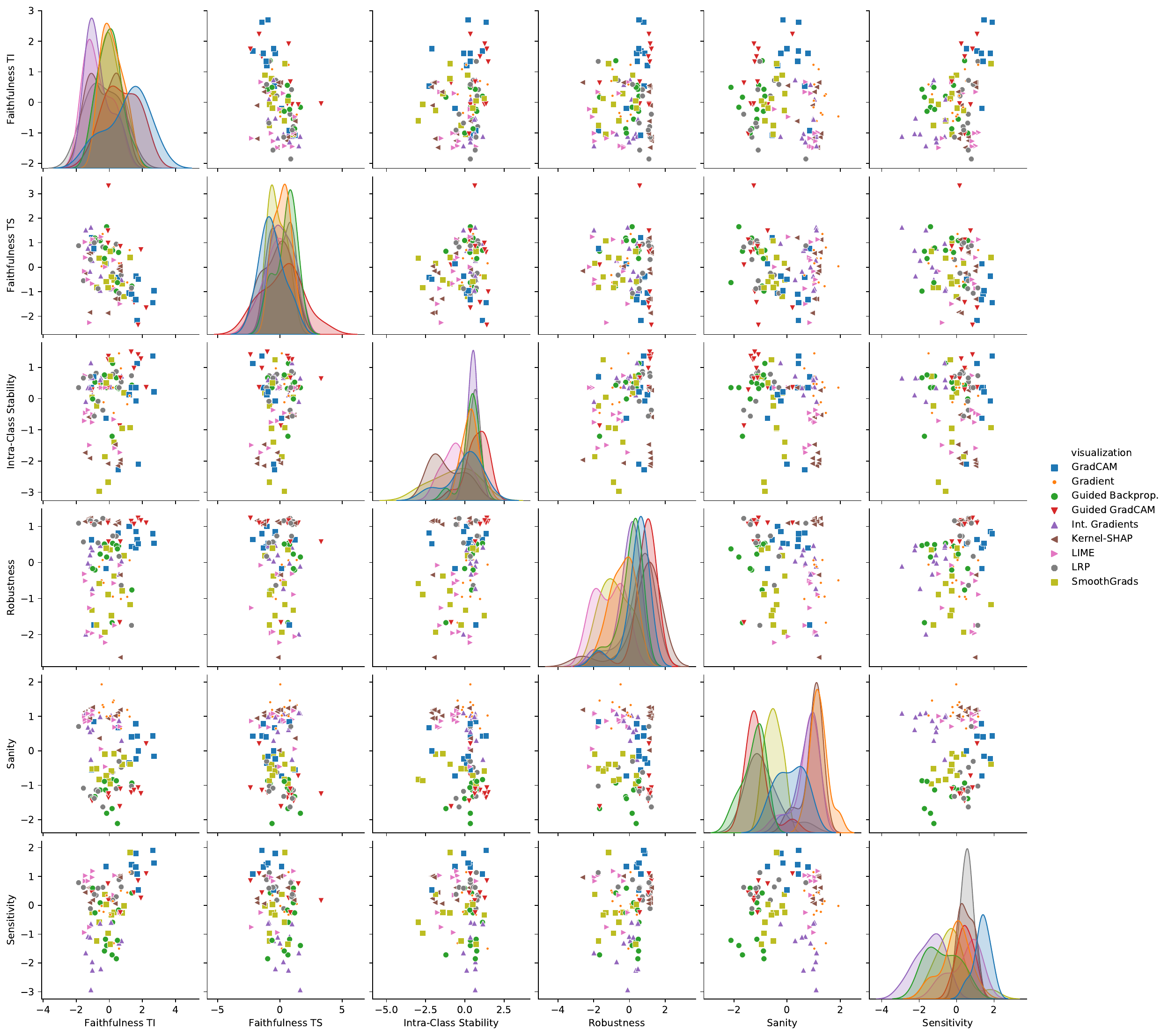}
     
    \caption{
    Pairplots of the correlations of metric scores with each other. No combination of metrics has a meaningful correlation with each other, proving that they provide independently helpful signals. This analysis is based on scores normalized for dataset bias, but the results also hold when this normalization is not done.
    }
    \label{appendix fig: correlations}%
\end{figure}

\begin{figure}[ht]
    \centering
     \includegraphics[width=1.0\textwidth, keepaspectratio]{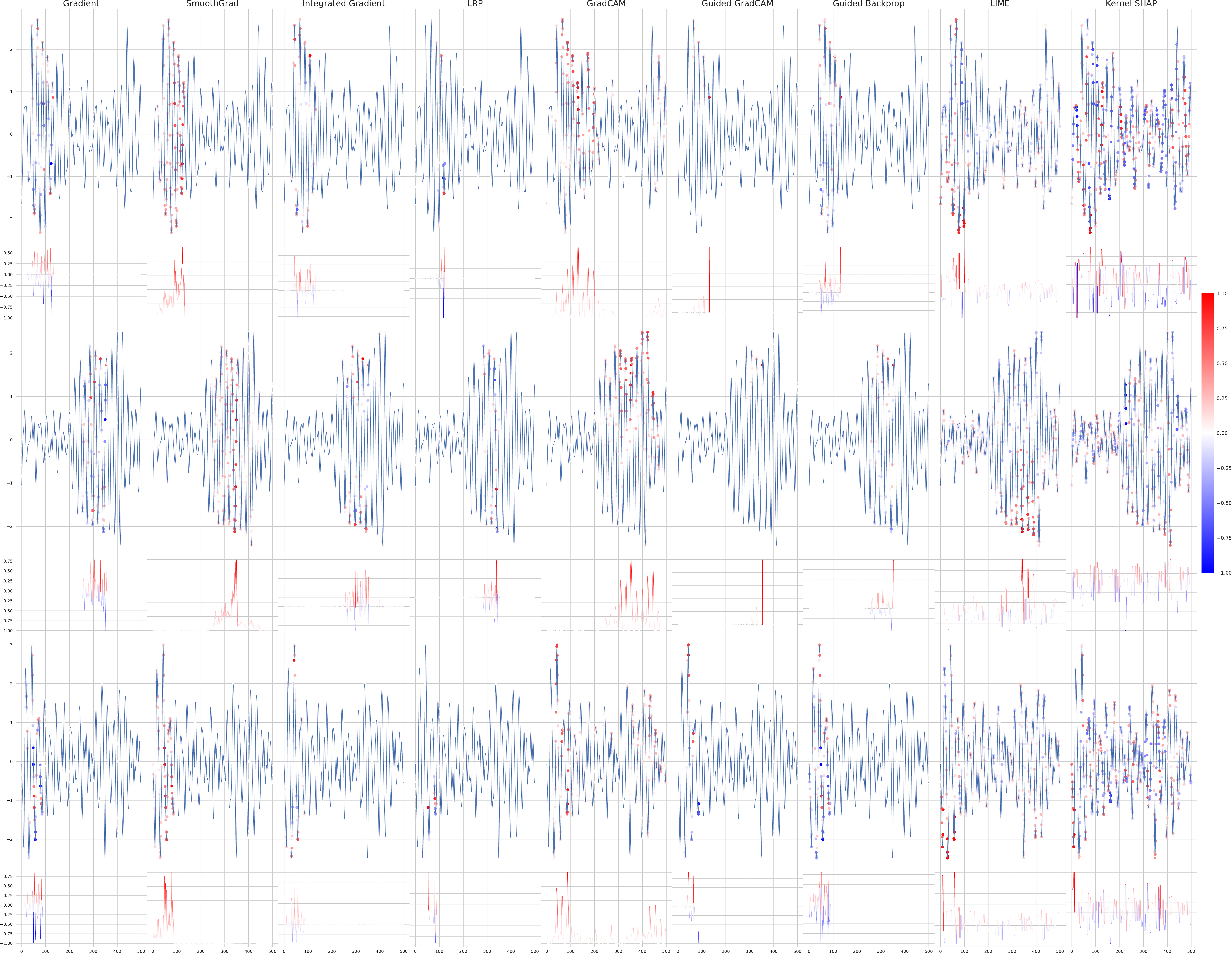}
    \caption{
    Examples of saliency maps for the TCN architecture on the FordB dataset for class 0.
    }
    \label{appendix fig: example class 0}%
\end{figure}

\begin{figure}[ht]
    \centering
     \includegraphics[width=1.0\textwidth, keepaspectratio]{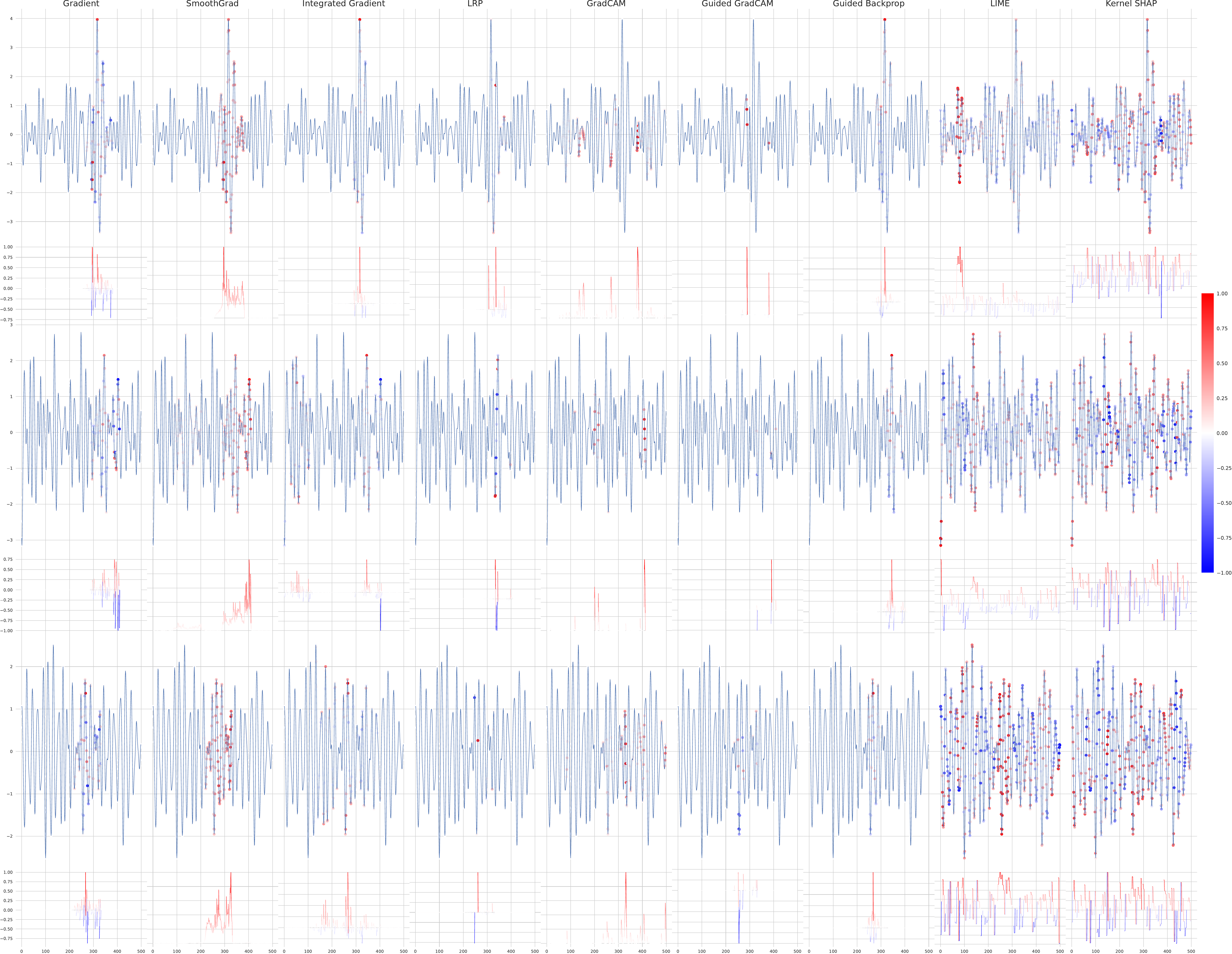}
    \caption{
    Examples of saliency maps for the TCN architecture on the FordB dataset for class 1.
    }
    \label{appendix fig: example class 1}%
\end{figure}

In Figure~\ref{fig: tool tracking full}, we show the saliency maps for all nine saliency methods for one instance of each of the four classes "Tightening", "Untightening", "ClockWise" and "CounterClockWise" for a TCN model on the Tool Tracking dataset.

\begin{sidewaystable}[ht]
        \begin{tabular}{lccccc}
        \toprule
        \textbf{Method} & \textbf{Faithfulness TI}& \textbf{Sanity} & \textbf{Sensitivity}  & \textbf{Robustness} & \textbf{Stability} \\
         & \textbf{$S_{\text{Faithfuless TI}}$}  & \textbf{$S_{\text{Sanity}}$} & \textbf{$S_{\text{Inter-Class. Sensitivity}}$} & \textbf{$S_{\text{Max Sensitivity}}$} & \textbf{$S_{\text{Intra-Class Stability}}$} \\
        \midrule
        GradCAM             & 0.85 & 0.07 & 0.18 & 0.09 & 8.12 \\
        Guided GradCAM      & 0.83 & 0.62 & 0.07 & 0.23 & 2.89 \\
        Gradients           & 0.65 & 0.30 & 0.06 & 0.23 & 4.57 \\
        Guided Backprop     & 0.51 & 0.45 & 0.19 & 0.16 & 3.51 \\
        Smooth Gradients    & 0.50 & 0.40 & 0.44 & 0.31 & 4.05 \\
        LRP                 & 0.43 & 0.54 & 0.01 & 0.24 & 3.99 \\
        Kernel Shap         & 0.38 & 0.02 & 0.02 & 6.17 & 9.03 \\
        LIME                & 0.30 & 0.06 & 0.12 & 4.09 & 6.14 \\
        Integrated Gradients & 0.5 & 0.15 & 0.03 & 0.23 & 4.08 \\
        \bottomrule
        \end{tabular}
        \caption{\label{tab:implementation_results}The five metrics for nine saliency methods on the Tool Tracking dataset for a TCN classifier model. In contrast to the recommendations, the scores in this table are not normalized over models and datasets. Instead, we show the scores as defined in Section~\ref{sec: methodology}, what facilitates their interpretation. Recall that higher $S_{\text{Faithfuless TI}}$ is better, lower $S_{\text{Sanity}}$ is better, higher $S_{\text{Inter-Class. Sensitivity}}$ is better, smaller $S_{\text{Max Sensitivity}}$ is better, and lower $S_{\text{Intra-Class Stability}}$ is better.}
\end{sidewaystable}


\end{appendices}


\bibliography{sample-base}

\end{document}